\documentclass[journal,twoside]{IEEEtran}
\ifCLASSINFOpdf
\else
\fi

\usepackage{amssymb}
\usepackage[colorlinks=False,linkcolor=blue]{hyperref}
\usepackage{amsmath}
\usepackage{graphicx}
\usepackage{caption}
\usepackage[dvipsnames]{xcolor}
\usepackage{subcaption}
\graphicspath{ {images/} }
\usepackage{float}
\usepackage{amsthm}
\usepackage{listings}
\usepackage{algorithm,algcompatible}
\makeatletter
\def\munderbar#1{\underline{\sbox\tw@{$#1$}\dp\tw@\z@\box\tw@}}
\makeatother

\usepackage{algorithm}
\usepackage{algorithmicx}
\usepackage{algpseudocode}
\usepackage{makecell}
\usepackage{booktabs}       
\usepackage{url}
\usepackage{multirow}
\usepackage{lipsum}
\usepackage{bbm}
\usepackage{soul}
\usepackage{mathtools}

\newtheorem{theorem}{Theorem}

\newtheorem{remark}{Remark}

\newtheorem{definition}{Definition}

\usepackage{lipsum}  
\usepackage{stmaryrd}

\usepackage{ upgreek }
\usepackage{textgreek}

\newcommand{\x}{\mathbf{x}}

\newcommand{\uu}{\mathbf{u}}
\newcommand{\s}{\mathbf{s}}

\newcommand{\R}{\mathbb{R}}
\newcommand{\Z}{\mathbb{Z}}

\usepackage{CJKutf8}
\usepackage{xcolor}
\usepackage{tikz}
\usetikzlibrary{shapes.geometric, arrows}
\usepackage{nomencl}
\makenomenclature

\begin{document}
%
\title{On the Comparability and Optimal Aggressiveness of the Adversarial Scenario-based \\ Safety Testing of Robots}

\author{Bowen Weng$^{1}$, Guillermo A. Castillo$^{1}$, Wei Zhang$^{2}$, and Ayonga Hereid$^{3}$
\thanks{This work was supported in part by National Natural Science Foundation of China under Grant No. 62073159, the Shenzhen Science and Technology Program under Grant No. JCYJ20200109141601708, and in part by the National Science Foundation under grant FRR-2144156. \textit{(Corresponding author: Wei Zhang.)}}
\thanks{$^{1}$Electrical and Computer Engineering, Ohio State University, Columbus, OH, USA;  {\tt\footnotesize \{weng.172, castillomartinez.2\}@osu.edu.}}
\thanks{$^{2}$SUSTech Institute of Robotics, Southern University of Science and Technology (SUSTech), China; {\tt\footnotesize zhangw3@sustech.edu.cn.}}
\thanks{$^{3}$Mechanical and Aerospace Engineering, Ohio State University, Columbus, OH, USA. {\tt\footnotesize hereid.1@osu.edu.}}%
}
\maketitle

\begin{abstract}
This paper studies the class of scenario-based safety testing algorithms in the black-box safety testing configuration. For algorithms sharing the same state-action set coverage with different sampling distributions, it is commonly believed that prioritizing the exploration of high-risk state-actions leads to a better sampling efficiency. Our proposal disputes the above intuition by introducing an impossibility theorem that provably shows all safety testing algorithms of the aforementioned difference perform equally well with the same expected sampling efficiency. Moreover, for testing algorithms covering different sets of state-actions, the sampling efficiency criterion is no longer applicable as different algorithms do not necessarily converge to the same termination condition. We then propose a testing aggressiveness definition based on the almost safe set concept along with an unbiased and efficient algorithm that compares the aggressiveness between testing algorithms. Empirical observations from the safety testing of bipedal locomotion controllers and vehicle decision-making modules are also presented to support the proposed theoretical implications and methodologies.
\end{abstract}


%
\IEEEpeerreviewmaketitle

\section{Introduction}\label{sec:introduction}
\IEEEPARstart{A} robot, such as an automated vehicle (AV) or a humanoid, is designed to operate in complex real-world functional domains interacting with other static and dynamic domain users, including other human participants. To ensure this interaction is safe (for the robot itself, for the domain environment, and for other domain users), the safety performance testing should be executed before, along with, and occasionally after the life span of any robotic product. This safety testing can serve various purposes, including falsification (identifying failure events), defect analysis (identifying causes of failure events), benchmark (safety performance comparison among various robots), certification, validation, and verification, to name a few. 

Across various safety testing purposes mentioned above, the testing system is typically a composition of the subject robot (SR) of unknown system dynamics and various behavioral modules (e.g. the perception, localization, decision-making, and control modules) subject to a certain group of testing actions and disturbances (e.g. the force one applies to the humanoid's torso or the driving behavior of other vehicles against the subject AV). The state of the testing system thus contains the typical dynamic states of the SR and various features from other testing participants and the environment. Note that the \emph{black-box} nature of the testing system implies that the state evolution supplied with a certain testing action is primarily unknown and mostly arbitrary. This configuration is particularly appreciated for safety regulatory, standardization, and rule-making purposes, which are also the primary applicable cases for the opinions and methods discussed in this paper.

Throughout this paper, we primarily focus on the scenario-based testing approach~\cite{riedmaier2020survey} which represents most, if not all, of the safety testing methods. In particular, given a certain testing system, a scenario-based safety testing algorithm repeatedly initiates runs of scenarios with each run specified by an initialization condition assigned to the testing system (e.g. position and velocity of the SR and other testing participants) along with a finite sequence of testing actions. The system state thus evolves following the testing system dynamics with the given testing actions. Note that the testing scenario is open-loop if the supplied testing actions are independent of the instantaneous state. Otherwise, one should expect a certain feedback testing policy with state-dependent admissible testing action space.

For each run of a scenario, the observed states are further collected and analyzed by a given safety cost function (e.g., the failure-or-non-failure check, various time-to-collision oriented metrics~\cite{lee1976theory,wishart2020driving} for AVs, and other lagging safety measures~\cite{manuele2009leading,weng2022finite}). The safety testing algorithm thus consecutively explores scenario runs until a certain termination condition is met related to the safety costs of all explored runs of scenarios. The termination condition is primarily determined by the safety evaluation purpose which can be of various forms. For example, given the failure-or-non-failure check as the single-run cost, the \emph{falsification}~\cite{riedmaier2020survey} algorithm terminates after encountering a certain number of failure runs of scenarios, and the observed risk metric~\cite{zhao2016accelerated,feng2021intelligent} terminates when the risk (the number of failure scenario runs divided by the total number of runs of scenarios) stabilizes with an acceptable tolerance of error.

The design of a scenario-based testing algorithm thus requires specifying (i) a state exploration strategy (i.e. how the initialization conditions are determined), (ii) an action exploration strategy (through open-loop action sampling or a certain feedback testing policy), (iii) the safety cost for each scenario run, and (iv) the cost-dependent termination condition of the algorithm. Algorithms are thus \emph{constructively different} with different designs of the above four components. As the safety cost and termination condition are primarily related to the safety testing purpose, this paper focuses on variants of the first two components, i.e. the state-action exploration mechanisms (assuming the same but arbitrarily selected cost and termination condition). This fundamentally covers most of the so-called \emph{adversarial} safety testing algorithms~\cite{ding2022survey} that prioritize the exploration of critical states, hostile testing actions and high-risk testing policies based on data-driven learning~\cite{feng2023dense, feng2021intelligent,koren2018adaptive,tuncali2018simulation,klischat2019generating,kuutti2020training,feng2020testing,li2020av,ding2021multimodal,innes2021automated}, analytical modeling~\cite{hussain2018lteinspector,capito2020modeled} and expert knowledge~\cite{najm2007pre,forkenbrock2015nhtsa,van2017euro,national2019traffic,winner2019pegasus}. It is commonly believed that an adversarial testing approach would accelerate the testing process~\cite{zhao2016accelerated,ding2022survey}, provide worst-case safety guarantee, enhance robustness and resilience of the SR against perturbations~\cite{madry2018towards}. 

\subsection{Problems and Challenges of Adversarial Testing}
However, the claimed success of various accelerated and adversarial scenario-based safety testing methods is mostly based on intuitions and empirical observations. To the best of the authors' knowledge, the formal definition of \emph{aggressiveness} (i.e. how adversarial the testing algorithm is) does not come with a common agreement in the safety testing literature and varies w.r.t. the construction of algorithms. Moreover, some of the fundamental properties remain unclear. In general, this paper is centered around two major properties.
\paragraph{Comparability} For scenario-based safety testing algorithms that are constructively different (at the state-action exploration strategies), how to make a fair comparison and claim one performs differently from another? 
\paragraph{Optimal Aggressiveness} In the case where the comparability does exist, what criterion should one use to justify the optimal testing algorithm and compare the testing aggressiveness? How do we provably and efficiently know that one safety testing method is more aggressive/adversarial than the other?

To formally study the above two performance properties, the constructive differences among algorithms w.r.t. the state-action exploration must be rigorously controlled. Two particular types of differences are considered in this paper. Type-I different algorithms share the same state-action set with different sampling distributions. The configuration is mostly applicable to the open-loop scenario design and represents a large group of adaptive/importance sampling based and some reinforcement learning based testing methods~\cite{klischat2019generating,li2020av,corso2019adaptive,ding2021multimodal}, as well as the expert-knowledge inspired concrete testing scenario design~\cite{najm2007pre,van2017euro}. Note that all Type-I different algorithms are coverage complete, i.e., as the number of explored runs of scenarios tends to infinity, the probability of visiting all admissible state-actions for each algorithm tends to one. On the other hand, Type-II different algorithms admit different feedback testing policies for testing action derivation~\cite{capito2020modeled,feng2021intelligent,wang2021interaction}, which then lead to different coverage of sets of state-actions. A more formal definition of the above two types of algorithms can be found in Section~\ref{sec:preliminaries_scenario}. 

For testing algorithms of the Type-I difference, one intuitive approach to justify the above two properties is through data efficiency. Existing work mostly assumes if the testing algorithms are constructed differently (i.e. sampling state-actions with different distributions or exploring state-actions in brute-force at different orders), then they also perform differently in terms of the data efficiency. That is, algorithms are different (comparable) at the number of runs of scenarios each one takes to reach the same termination condition (for an arbitrarily given cost and termination definition) if the state-actions are explored following different sampling distributions. However, as we should address in detail later in Section~\ref{sec:comparability}, the aforementioned assumption is not always correct, especially under the black-box testing configuration. As a result of the non-existence of the comparability property, the optimal aggressiveness is not applicable either.

Moreover, as one considers the class of testing algorithms of Type-II difference, one directly validates the comparability property (i.e. algorithms perform differently) given the state-action coverage among algorithms are diffident. However, one can no longer rely on the aforementioned notion of sampling efficiency to justify the optimal performance as testing algorithms of different feedback testing policies do not necessarily satisfy the same termination condition. For algorithms of the above nature, existing methods tend to draw equivalent correlations between the aggressiveness with the magnitude of testing actions (e.g. high-speed~\cite{barcelo2003safety}, large acceleration and jerk values~\cite{bellem2018comfort} imply aggressive behavior of vehicles) and the observed failure risk~\cite{feng2021intelligent,li2020av}. This can be inaccurate and unfair as the basic \emph{transitive property} is not necessarily satisfied. That is, a robot tested safe by the more adversarial testing algorithm is not always safe against the less adversarial one with the aggressiveness deemed by one of the aforementioned equivalences. Some practical examples of this deficiency are presented in Section~\ref{sec:case}. 

In general, existing studies of the above two classes of scenario-based safety testing algorithms regarding the two major properties (comparability and optimal aggressiveness) have been primarily taking a data-driven approach with empirical observations from testing a particular SR within a concrete testing system. This paper is inspired to rethink and re-evaluate the above testing algorithms in a more formal manner within the black-box configuration. The main contributions are further summarized w.r.t. the two major properties of interest separately as follows.

\subsection{Main Contributions}
\textbf{Comparability}: We present, to the best of our knowledge, the first impossibility theorem in the safety testing regime. Through an adaptation of the No-Free-Lunch (NFL) theorem from the optimization and machine learning regime~\cite{wolpert1997no,wolpert2021important}, we formally prove that all scenario-based testing algorithms of Type-I difference perform equally well in terms of the expected sampling efficiency over any definition of safety cost and termination condition in the black-box testing configuration. This provides fundamental insights to some of the empirically observed deficiencies in the adversarial learning regime~\cite{lechner2021adversarial} and safety testing field~\cite{hauer2020re,capito2020modeled} where a seemingly adversarial testing strategy for one robot does not necessarily hold the same level of testing aggressiveness for another. 

\textbf{Optimal Aggressiveness}: For testing algorithms of different feedback testing policies (i.e. Type-II difference), the proposed impossibility theorem no longer applies. The comparability property is thus possible and the algorithms perform differently. However, to fairly claim that one algorithm is more adversarial than the other, one has to formally define the testing algorithm aggressiveness. Different from the aforementioned techniques, we then propose a set-based metric to fulfill the desired definition of optimal aggressiveness by specifying (i) the unbiased approximation of the particular state(-action) set within which the SR is probably safe against the given feedback testing policy and (ii) the probability for the claim to hold. The proposal is, intuitively and provably, a better alternative than the aforementioned action magnitude related and observed risk based aggressiveness measures. It is an adaptation of the $\epsilon\delta$-almost safe set method previously proposed for safety performance justification of AVs~\cite{weng2021towards,weng2022formal} and bipedal robots~\cite{weng2022leeged}. Finally, we also propose a data-driven set-based methodology that formally compares the testing algorithms efficiently by the proposed definition of aggressiveness. 
\subsection{Construction}
The overall construction of the paper is as follows. Section~\ref{sec:preliminaries_scenario} reviews the basics of the scenario-based safety testing algorithm and formulates the two types of constructively different testing algorithms. The set-based safety evaluation approach is reviewed in Section~\ref{sec:preliminaries_metric}. Section~\ref{sec:comparability} addresses the comparability property of Type-I different algorithms as mentioned above. Section~\ref{sec:optimality} discusses the optimal aggressiveness property w.r.t. the Type-II different algorithms. In Section~\ref{sec:case}, a series of empirical observations are presented with safety testing applications with bipedal robot locomotion controllers and vehicle decision-making modules. The observations support the various theoretical insights discussed in Section~\ref{sec:comparability} and Section~\ref{sec:optimality}. Conclusions and future work of interest are further discussed in Section~\ref{sec:conclusion}.

\textbf{Notation: } The set of real and positive real numbers are denoted by $\R$ and $\R_{>0}$, respectively. $\Z$ denotes the set of all positive integers and $\Z_N=\{1,\ldots,N\}$. $|X|$ is the cardinality of the set $X$, e.g., for a finite set $D$, $|D|$ denotes the total number of points in $D$. $|\x|$ can also denote the absolute value for some $\x \in \R^n$. $\llbracket i, j \rrbracket$ denotes the Kronecker delta function as 
\begin{equation}\label{eq:kdf}
    \llbracket i, j \rrbracket = 
    \begin{cases*}
    1 & if $i = j$ \\
    0 & otherwise
  \end{cases*}.
\end{equation}
Some commonly adopted acronyms are also adopted including i.i.d. (independent and identically distributed), w.r.t. (with respect to), and w.l.o.g. (without loss of generality). One can refer to the Appendix~\ref{apx:notation} for the nomenclature of other important notations used in this paper.

\section{Preliminaries}\label{sec:preliminaries}
We first revisit the scenario-based test and formulate the class of black-box safety testing algorithms. The set-based safety metric is also reviewed in Section~\ref{sec:preliminaries_metric}. 

\subsection{Scenario-based Test}\label{sec:preliminaries_scenario}
A testing system involves a SR (subject robot) with a certain subject dynamics controlled by a combination of various behavioral modules along with other dynamic participants and environmental factors. A testing operator controls the controllable inputs such as the environmental disturbances (e.g. external forces) and the behavior of other testing participants (e.g. the behavior of other traffic vehicles and pedestrian). Observed and extracted states are then collected through a certain discrete-time data acquisition system. This leads to the motion dynamics of the following form.
\begin{equation}\label{eq:ctrl-sys}
    \s(t+1) = f_s(\s(t), \uu(t); \omega(t)).
\end{equation}
The state $\s \in S \subset \R^n$ and $S$ denotes a finite but possibly large set of states. The action $\uu \in U$ represents the control inputs of the testing system. $U$ is thus the (finite) testing action space. It is occasionally denoted as $U(\s)$ if the admissible testing action set is state-dependent on $\s \in S$. $\omega \in W$ and $W$ denotes the set of environmental disturbances and uncertainties. The discrete finite set assumption is a practically appropriate configuration as practical testing execution and data acquisition are both with digital equipment of finite-bit precision. Let $F_s=S^{S\times U \times W}$ denote the space of all possible dynamic transitions within the specified state-action set (e.g. all possible combinations of mechanical designs of bipedal robots, perception algorithms, decision-making strategies, and locomotion controllers). Each $f_s \in F_s$ denotes a specific testing system with a set of fixed choices of hyper-parameters and configurations. In practice, the same robots with different versions of the software stack contribute to different $f_s$ (testing systems), robots of different makes and models also lead to different $f_s$. Differences among the different $f_s$ are partially known by the designer and manufacturer of each individual robot, but are in general unknown from the testing and evaluation perspective. This fundamentally leads to the \emph{black-box} nature of the safety testing which admits the following property.
\begin{remark}\label{rmk:bbox}
    In the black-box testing configuration, the testing system dynamics $f_s\in F_s$ is unknown and arbitrary. Without known prior of any system details, $f_s$ is uniformly distributed in $F_s$.
\end{remark}
Note that the above remark formally defines the intuitive notion of ``a robot can do anything" within the selected functional domain. In the existing literature, Corso et al.~\cite{corso2020survey} specify the notion as ``black-box techniques do not assume that the internals of the system are known" and ``they consider a general mapping from input to output that can be sampled." This conceptual description clearly aligns with the proposed Remark~\ref{rmk:bbox}.

Let $O \subseteq S$ be a set of states that are of primary concern for a certain functionality or work domain of the robot, referred to as its operational state space (OSS)~\cite{weng2022finite}. The space of all black-box testing systems $F_s$ then admits the form as $F_s = O^{O \times U \times W}$. Let $\mathcal{C}$ be a set of failure states such as collisions and falling-over. Note that OSS and $\mathcal{C}$ are non-unique in general (as a robot can be expected to achieve various desired functionalities and experience different types of failure events). One can refer to~\cite{weng2022finite} for examples of OSS designs in the AV field and~\cite{weng2022leeged} for OSS examples with bipedal and humanoid robots. One other terminology closely related to OSS is the \emph{Operational Design Domain} (ODD)~\cite{gyllenhammar2020towards} adopted to characterize the work domain within which the SR is expected to be safe for the tested functionality. The ODD is thus a subset of the OSS.

A scenario-based test of a certain SR in the testing system $f_s$ using a sequence of testing actions thus collects and analyzes a group of sampled trajectories that characterizes the evolution of the state variables of system~\eqref{eq:ctrl-sys} within the OSS of concern. Formally speaking, a scenario is a function of the form $\sigma(k): \Z_k \rightarrow O$. 
Note that the state evolution within a scenario starts and remains within the studied OSS unless a failure event occurs. That is, if a test scenario reaches states outside the studied OSS at a certain time $t \in \Z_{\geq1}$, we consider $\sigma(t+t')=\sigma(t-1)$ for all $t' \in \Z$ until the trajectory gets back to $O$. Note that $\sigma(t) \notin O$ does not necessarily indicate an unsafe outcome. It could also represent the case where the failure concern is unnecessary as one shall see later in the discussion related to Fig.~\ref{fig:example}. Such an enforced condition is not a precise description of the motion revealed by the testing system dynamics, but it copes well with the limited data acquisition capability in practice, and it does not affect the accuracy or the unbiasedness of the proposed safety evaluation outcomes presented in this paper. In practice, one could also make $O$ sufficiently large, but the testing efficiency will be jeopardized with very little added value to the safety performance testing tasks. Finally, if a test scenario indeed encounters a failure state at a certain time $t$ (i.e. $\sigma(t)\in\mathcal{C}$), we then have $\sigma(t+t')\in\mathcal{C}$ for all $t'\in\Z_{k-t}$.

The sequence of actions that controls the evolution of states within each scenario can be determined in two ways based upon the state-dependency. The open-loop approach explores all actions $\uu \in U$ following either a certain concrete order (brute-force) or a certain sampling distribution. The state-dependent method typically admits a specific feedback testing policy $\pi$ as
\begin{equation}\label{eq:test_pi}
    \uu = \pi(\s), 
\end{equation} 
We then have a composed testing system dynamics of 
\begin{equation}\label{eq:sys}
    \s(t+1) = f(\s(t); \omega(t))
\end{equation}
by replacing $\uu$ in~\eqref{eq:ctrl-sys} with \eqref{eq:test_pi}. In practice, such a concrete testing policy can emulate the exact feedback testing strategy the SR is expected to encounter in the given domain (e.g. the real-world statistical naturalistic driving policy against an AV as the SR). The testing policy can also test the SR more aggressively by prioritizing the high-risk actions dependent upon the given states.

For a scenario-based test, the choice between exploring all actions in $U$ in the open-loop fashion and following a feedback policy $\pi$ is primarily determined by the safety evaluation purpose. For complete falsification (finding some or all states and actions that lead to failure events), the scenario-based testing covering the complete state-action set is commonly adopted. On the other hand, some application require the robot to operate in a specific functional domain against a specific testing strategy, such as the AV operating against human drivers and the humanoid walking in the crowd of pedestrians. It is thus a natural decision to adopt a concrete testing policy that emulates the naturalistic driving policy or the crowd motion behavior. Other factors such as the system dynamics nature and ethical concerns could also affect the testing action propagation scheme as we will see in Section~\ref{sec:case}.

Regardless of how testing actions are generated, a \textit{run of a scenario}, $\mathcal{R}_{\sigma}(\s_0, k)$, is defined as a sequence of acquired states $\{\sigma(i)\}_{i=1,\ldots,k}$ and $\sigma(0)=\s_0$. Note that given fixed $\s_0$ and $k$, the run of a scenario is not necessarily unique with the presence of disturbances and uncertainties in~\eqref{eq:ctrl-sys}. W.l.o.g, let all scenarios be defined over the same time domain. With a little abuse of notation, $\mathcal{R}_{\sigma}(\s_0, k)$ is simplified as $\mathcal{R}(\s_0)$ for the remainder of this paper. The testing action sequence associated with the run of a scenario is also simplified as $\bar{\uu} = \{\uu_j\}_{j\in\Z_{k}}$.

Let $c: S^k \rightarrow G$ be a cost function that takes a run of a scenario $\mathcal{R}(\s_0)$ as the input. Note that $g \in G$ denotes the object that characterizes a certain safety property of the SR revealed through a particular run of a scenario. Such an object can be as simple as a Boolean value (e.g. the failure-or-non-failure check as $\llbracket \mathcal{R}(\s_0) \cap \mathcal{C}, \emptyset \rrbracket$), but can also take complex forms such as the scenario run itself. As the inputs are finite, the set $G$, while sometimes can be quite large, is also finite. Moreover, the term \emph{sample} is used to refer to all inputs (e.g. $\s_0$ and testing action sequence) and outputs (e.g. the collected states and the cost) related to a single run of a scenario.

Consider $m\in \Z$ samples with a certain SR within the testing system $f_s$, let 
\begin{equation}
    g_m = \left[ c(\mathcal{R}(\s_0^1)), \ldots, c(\mathcal{R}(\s_0^m)) \right]
\end{equation}
be the sequence of safety costs obtained from the aforementioned $m$ scenario runs.

A scenario-based safety testing algorithm is thus formulated as 
\begin{equation}\label{eq:eval}
    \mathcal{TE}: G^m \rightarrow S \times U^{k}.
\end{equation}
That is, the algorithm consecutively determines how to collect the next run of a scenario (i.e. the initialization state $\s_0$ and the sequence of testing actions $\bar{\uu}$) based on the previously obtained $m$ costs. Note that some algorithms determine each scenario run independently of the historical exploration, which can be viewed as a special case of~\eqref{eq:eval}. 

The termination condition of the algorithm takes various forms such as if a certain safety cost (e.g. one failure run of a scenario) or a particular group of costs have been collected. W.l.o.g., consider the termination function $\mathcal{T}$ taking the instantaneous sequential costs $g_m$ as the input and the algorithm $\mathcal{TE}$ always terminates if the obtained sequential costs $g_m$ satisfies $\mathcal{T}(g_m)=\texttt{True}$. 

In summary, the class of scenario-based safety testing algorithms admits the basic form as shown in Algorithm~\ref{alg:te_u} and Algorithm~\ref{alg:te_pi}. In particular, $\mathcal{TE}^{u}$ denotes the class of safety testing algorithms with open-loop testing scenarios that explore all initialization states and testing action trajectories in a coupled space of $S \times U^{k}$ through a selected order or sampled sequence. $\mathcal{TE}^{\pi}$ chooses to follow a state-dependent way of testing action derivation through a certain feedback testing policy $\pi$. In both algorithms, the testing system $f_s\in F_s$ is implicitly given as a black-box. 

Note that we assume all sampled runs of scenarios are different, i.e. there does not exist any pair of runs of scenarios sharing the identical combination of state trajectory, testing action sequence, and safety cost. This is certainly feasible in practice as one simply ignores identical outcomes given no added safety information can be gained from the duplicated tests. As a result, exploring all state-actions at different orders and sampling state-actions following different distributions are essentially equivalent.

\begin{algorithm}[H]
    \begin{algorithmic}[1]
    \State {\bf Given: $\mathcal{T}(\cdot)$, $c(\cdot)$, $O \subset S$ and $U$}
    \State {\bf Input: $g_m$}
    \State {{\bf While} $\mathcal{T}(g_m)=\texttt{False}$:}
    \State {\ \ \ \ Determine $\s_0 \in O$}
    \State {\ \ \ \ Determine the testing action sequence $\bar{\uu} \in U^{k}$}
    \State {\ \ \ \ Collect $\mathcal{R}(\s_0)$ with $\bar{\uu}$ through $f_s$~\eqref{eq:ctrl-sys}}
    \State {\ \ \ \ $g_m$.\texttt{add}($c(\mathcal{R}(\s_0))$) }
    \State {\ \ \ \ $m += 1$}
    \State {\ \ \ \ $\mathcal{TE}^u(g_m)$}
    \end{algorithmic}
    \caption{$\mathcal{TE}^u(g_m)$} \label{alg:te_u}
\end{algorithm}

\begin{algorithm}[H]
    \begin{algorithmic}[1]
    \State {\bf Given: $\mathcal{T}(\cdot)$, $c(\cdot)$, $O \subset S$ and $\pi(\cdot)$}
    \State {\bf Input: $g_m$}
    \State {{\bf While} $\mathcal{T}(g_m)=\texttt{False}$:}
    \State {\ \ \ \ Determine $\s_0 \in O$}
    \State {\ \ \ \ Collect $\mathcal{R}(\s_0)$ and $\bar{\uu}$ with $\pi(\cdot)$ composed through~\eqref{eq:ctrl-sys}}
    \State {\ \ \ \ $g_m$.\texttt{add}($c(\mathcal{R}(\s_0))$)  }
    \State {\ \ \ \ $m += 1$}
    \State {\ \ \ \ $\mathcal{TE}^{\pi}(g_m)$}
    \end{algorithmic}
    \caption{$\mathcal{TE}^{\pi}(g_m)$} \label{alg:te_pi}
\end{algorithm}

Finally, the general notion of ``different safety testing algorithms" can occur in various ways. Within this paper, we are primarily interested in two types of commonly observed differences defined as follows.
\begin{definition}\label{def:diff}
    Let $\mathcal{TE}_1$ and $\mathcal{TE}_2$ belong to the same sub-category of $\mathcal{TE}$ (either $\mathcal{TE}^{u}$ or $\mathcal{TE}^{\pi}$). The two safety evaluation algorithms are considered different for one of the following two types.
    \begin{enumerate}
        \item \textbf{Type I}: $\mathcal{TE}^u_1$ and $\mathcal{TE}^u_2$ share the same configuration (i.e. the same state-action space, the same cost, and the same termination condition) except for the different state-action exploration order or different probability distributions of states and actions over the same sample space.
        \item \textbf{Type II}: $\mathcal{TE}_1$ and $\mathcal{TE}_2$ admit different sets of testing actions $U^1(\s)\neq U^2(\s)$ for some or all $\s \in O$ (or different testing policies $\pi^1 \neq \pi^2$) with all other configurations remain identical.
    \end{enumerate}
\end{definition}

Note that, the above algorithmic difference is defined w.r.t. the construction of the algorithm. Whereas the comparability and optimal aggressiveness as we should address later in this paper is concerned with the algorithms' performance. In general, the constructive difference does not necessarily imply performance difference among algorithms. Moreover, safety testing algorithms may also exhibit other types of differences or a combination of the above two. One particular example is the combination of the open-loop exploration of testing parameters associated with a certain parameterized feedback testing policy. This technically leads to the Type-I difference as the feedback testing policy can be viewed as part of $f_s$, yet it also makes the testing system not necessarily a black-box as we should see in Section~\ref{sec:case_cassie}. More other variants are beyond the scope of this paper.

For all algorithms of Type-I difference, the same termination condition is always achievable as the algorithm is coverage complete (i.e. as the number of executed runs of scenarios tends to infinity, the probability of encountering all reachable runs of scenarios tends to one). However, given the same cost design and termination condition, the number of runs of scenarios one takes to reach the termination point varies as the explicit order in which the states and actions are explored varies. Intuitively, a more adversarial testing strategy is expected to take a smaller number of samples for all testing systems $f_s \in F_s$. Hence the adversarial testing algorithm is equivalent with the accelerated testing algorithm. To argue that the safety evaluation algorithms of Type-I difference are comparable, it requires to show that one algorithm can be faster than another in reaching a certain termination condition in the black-box testing configuration against all $f_s \in F_s$. This inspires the comparability discussion in Section~\ref{sec:comparability}.

Moreover, algorithms of Type-II difference are not guaranteed to share the same $G$ supplied with the same cost function design as they do not share the same coverage of actions or runs of scenarios. The comparison among algorithms in this case is essentially a comparison among different $U$ (testing action spaces) or $\pi$ (feed-back testing policies). One thus requires a different methodology to define the optimal performance of an adversarial safety testing algorithm. The following review of the basics of the set-based safety metric will facilitate our proposal regarding the optimal aggressiveness in Section~\ref{sec:optimality}.

\subsection{Set-based Safety Performance Characterization}\label{sec:preliminaries_metric}
The safety performance characterization metric, or simply a safety metric, is typically adopted to justify the safety performance of a given SR against a certain testing approach. Formally speaking, consider a certain OSS $O \subset S$ and $f_s\in F_s$ associated with a certain SR, and a certain state-dependent testing action set $U(\s)$ or a certain testing policy $\pi$. A safety metric follows the primary task to characterize how the SR performs, in terms of safety, against all testing actions in $U(\s)$ at all states $\s \in O$ or against the testing actions dictated by $\pi$ in $O$. On the other hand, one can also flip the side and use the same safety metric to measure how various selections of feedback testing policies perform against the same set of testing system dynamics. This makes the following discussion useful for the testing algorithm aggressiveness comparison as we will introduce later.

We start from the algorithm $\mathcal{TE}^{\pi}$ with a certain testing policy $\pi$ as this is the relatively better studied form by the referred literature. The definition of the $\epsilon$-almost safe set~\cite{weng2021towards} taking consideration of the system randomness is presented as follows. 
\begin{definition}\label{def:epsilon-almost-ss}
    Given $\epsilon \in (0,1]$, a set $\Phi \subseteq O$. For any $\s_0 \in \Phi$, the run of a scenario $\mathcal{R}(\s_0)$ is collected following a certain testing policy $\pi$ through $f_s \in F_s$. $\Phi$ is the $\epsilon$-almost safe set for the system $f_s$ against $\pi$ if $\Phi \cap \mathcal{C} = \emptyset$ and
    \begin{equation}\label{eq:epsilon-almost-ss}
        \mathbb{P}\Big(\big\{\s_0 \in \Phi: \mathcal{R}(\s_0) \setminus \Phi \neq \emptyset \ \big\}\Big)\leq \epsilon.
    \end{equation}
\end{definition}
That is, the robot is almost safe against $\pi$ in $O$ following $f_s$ except for an arbitrarily small subset dictated by the probability coefficient $\epsilon$. 
Such an obtained $\epsilon$-almost safe set and its various characteristics (e.g. coverage, cardinality, and shape) are safety performance features. The comparison among various SRs can be further justified by comparing the obtained $\epsilon$-almost safe sets in $O$ against the same testing policy. Note that the above definition also fundamentally implies $\Phi$ being a robustly forward invariant set~\cite{blanchini1999set} for~\eqref{eq:sys}. 
 
Moreover, the validation of the $\epsilon$-almost safe set by Definition~\ref{def:epsilon-almost-ss} requires consecutively observing a sufficient number of runs of scenarios remaining inside the set, which is formally justified through the following theorem adapted from~\cite{weng2021formal}.
\begin{theorem}\label{thm:validation}
    Given $f_s\in F_s$, $\epsilon \in (0,1]$, $\beta \in (0,1]$, $\Phi \subseteq O$. Consider $N$ runs of scenarios, $\{\mathcal{R}(\s_0^i)\}_{i=1,\ldots,N}$, with the state initialization of each run being i.i.d. w.r.t. the underlying distribution on $\Phi$ and the testing actions generated through the testing policy $\pi$. The set $\Phi$ is the $\epsilon$-almost safe set for the system $f_s$ against $\pi$ with confidence level at least $1-\beta$ if
    \begin{equation}
        \bigcup_{i=1}^N \mathcal{R}(\s_0^i) \subseteq \Phi, \bigcup_{i=1}^N \mathcal{R}(\s_0^i)  \cap \mathcal{C}=\emptyset,
    \end{equation}
    and
    \begin{equation}\label{eq:validation_N}
        N \geq \frac{\ln{\beta}}{\ln{(1-\epsilon)}}.
    \end{equation}
\end{theorem}
The above theorem immediately leads to a safety evaluation algorithm $\mathcal{TE}^{\pi}$ admitting the same steps described in Algorithm~\ref{alg:te_pi}. Given the initial OSS $O=\Phi$, the algorithm consecutively samples runs of scenarios starting from $\s_0 \in \Phi$ and propagates through $f_s$ against the testing policy $\pi$. The cost function admits the aforementioned failure-or-non-failure cost design. At the $m$-th iteration, if the executed run of scenario ends up in $\mathcal{C}$, the algorithm is terminated and the SR is unsafe in the given set $\Phi$. Otherwise, $\Phi$ is an $\epsilon_m$-almost safe set with confidence level at least $1-\beta$ and $\epsilon_m = 1-\exp{(\frac{\ln \beta}{m})}$ (a direct derivation from~\eqref{eq:validation_N}). The algorithm then terminates with a sufficiently small $\epsilon_m$ deemed by the testing operator. 

In practice, as one rarely knows the appropriate candidate set that is indeed the $\epsilon$-almost safe set with a certain confidence level justified by Theorem~\ref{thm:validation}, the above validation commonly ends up with a falsified outcome. The class of \emph{optimal safe domain quantification} algorithms is thus inspired to characterize the largest subset of the initial OSS that is $\epsilon$-almost safe. Given an arbitrary candidate set of states, the optimal safe domain quantification algorithm keeps exploring runs of scenarios, modifying the candidate set, until the exploration stabilizes with $\frac{\ln{\beta}}{\ln{(1-\epsilon)}}$ consecutive runs of scenario remaining in the largest candidate set by Theorem~\ref{thm:validation} given $\epsilon$ and $\beta$. 

\begin{figure*}
    \centering
    \includegraphics[width=\textwidth]{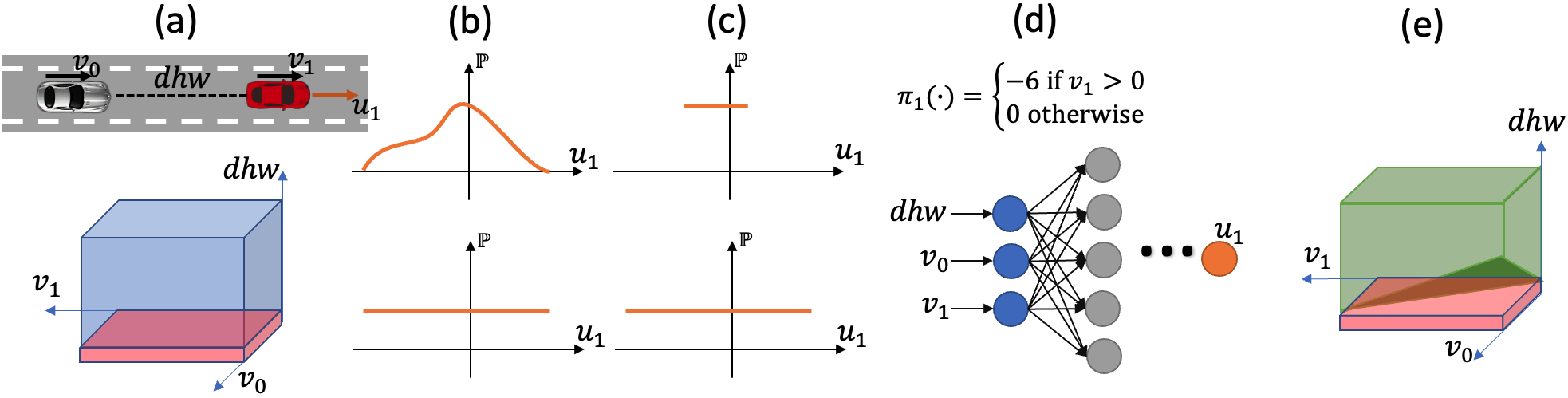}
    \caption{An example for some of the definitions introduced in Section~\ref{sec:preliminaries}: (a) a standard lead-vehicle following testing system and its corresponding OSS design, (b) an example of the Type-I difference, (c) an example of the Type-II difference w.r.t. $\mathcal{TE}^u$, (d) an example of the Type-II difference w.r.t. $\mathcal{TE}^{\pi}$, (e) a conceptual illustration of the almost safe set.}
    \label{fig:example}
    \vspace{-5mm}
\end{figure*}

Note that many of the set-based safety evaluation algorithms, especially the class of quantification algorithms, can be computational inefficient due to the extremely large cardinality of $O$. An immediate alternative is to characterize the $\epsilon$-almost safe set in a controlled less accurate manner without sacrificing the impartiality. One such example considered by previous work~\cite{weng2021formal} is through the extended $\delta$-covering set defined as follows, which is simplified as the $\delta$-covering set throughout this paper. 
\begin{definition}\label{def:ext-delta-cover}
    Given a certain set $O \subset \R^n, n\in \Z$ and $\delta \in \R_{>0}^n$, let $\mathcal{N}_{\delta}(\s)$ be the $\delta$-neighbourhood of $\s$, i.e., 
    \begin{equation}\label{eq:ext-delta-neib}
        \forall \s'\in \mathcal{N}_{\delta}(\s), |\s-\s'| \leq \delta.
    \end{equation}
    We claim that $\Phi_{\delta}^{O}$ is an $\delta$-covering set of $O$ if for some $k \in \Z$ and $\s_i\in O, i=1,\ldots,k$, we have
    \begin{equation}
        \Phi_{\delta}^{O}\!=\!\bigcup_{i\in\{1,\ldots,k\}} \mathcal{N}_{\delta}(\s_i) \supseteq O \text{ and } \Phi_{s}^{O}\!=\!\{\s_i\}_{i\in\{1,\ldots,k\}} \subseteq O.
    \end{equation}
    Furthermore, $\Phi_{s}^{O}$ are centroids of $\Phi_{\delta}^{O}$.
\end{definition}
One can refer to Fig.~2 in~\cite{weng2021towards} for a special illustrative example of the above definition on $\R^2$ with both entries of $\delta$ share the same value.
Note that the inequality in \eqref{eq:ext-delta-neib} is element-wise. Moreover, the $\delta$-covering set and the set of centroids are not necessarily required to be affiliated with an explicit set to be covered (i.e. $O$). The notions of $\Phi_{\delta}$ and $\Phi_{s}$ are thus adopted. They are primarily for cases where the $\delta$-covering set is defined over an implicit target $O'$ satisfying $\Phi_{s} \subseteq O' \subseteq \Phi_{\delta}$.

The main advantage of Definition~\ref{def:ext-delta-cover} is that the set of centroids $\Phi_{s}^{O}$ forms a representative subset of $O$ while ensuing uniform coverage through the $\delta$-neighbourhood. Taking the validation algorithm for example, if incorporated with the $\epsilon$-almost safe set, one can simplify the process from exploring all $\s \in O$ to a sufficient validation of runs of scenarios initialized from $\Phi_{s}^{O}$ and never leave $\Phi_{\delta}^{O}$. This leads to a series of algorithms relying on the so-called $\epsilon\delta$-almost safe set~\cite{weng2021towards} defined as follows.
\begin{definition}\label{def:epsilondelta-almost-ss}
    Given an extended $\delta$-covering set $\Phi_{\delta} \subset S$ with centroids $\Phi_s \subset \Phi_{\delta}$, $\delta \in \R_{\geq0}^N$. Let $\epsilon \in (0,1]$. $\Phi_{\delta}$ is the $\epsilon\delta$-almost safe set for the system $f_s$ against $\pi$ if $\Phi_{\delta} \cap \mathcal{C} = \emptyset$ and
    \begin{equation}\label{eq:epsilondelta-almost-ss}
        \mathbb{P}\Big(\big\{\s_0 \in \Phi_s: \mathcal{R}(\s_0) \setminus \Phi_{\delta} \neq \emptyset \ \big\}\Big)\leq \epsilon.
    \end{equation}
\end{definition}
Moreover, given
\begin{equation}
    O = \lim_{\delta\rightarrow 0}\Phi_{\delta}^{O} = \lim_{\delta\rightarrow 0}\Phi_{s}^{O},
\end{equation}
one ensures the \emph{resolution completeness} property of any safe domain validation \& quantification algorithms incorporating Definition~\ref{def:epsilondelta-almost-ss}. As $\delta$ tends to zero, one is guaranteed to validate or recover the complete almost safe set. One can refer to the previous literature and open-source code on more concrete examples of existing safe domain validation and quantification algorithms~\cite{weng2021towards,weng2021formal,weng2022leeged,sdq_tools}.

We conclude this section by extending the $\epsilon$-almost safe set obtained for algorithm $\mathcal{TE}^{\pi}$ against the testing policy $\pi$ to $\mathcal{TE}^{u}$ against the testing action space. Fundamentally speaking, the robustly forward invariant set is further extended to the robustly controlled forward invariant set~\cite{blanchini1999set}, leading to the $\epsilon$-almost controlled safe set defined as follows.
\begin{definition}\label{def:epsilon-almost-ssu}
    Given $\epsilon \in (0,1]$, a state set $\Phi \subseteq O$ and the set of admissible actions for each $\s \in \Phi$ denoted as $U(\s)$. $\Phi$ is the $\epsilon$-almost controlled safe set on $U(\s), \forall \s \in \Phi$ for the system $f_s$ if $\Phi \cap \mathcal{C} = \emptyset$ and
    \begin{equation}\label{eq:epsilon-almost-ssu}
        \mathbb{P}\Big(\big\{\s_0 \in \Phi, \bar{\uu}\in \cup_{i\in\Z_{k}} U(\s_i) : \mathcal{R}(\s_0) \setminus \Phi \neq \emptyset \ \big\}\Big)\leq \epsilon.
    \end{equation}
\end{definition}
One can also consider a special case of the above definition with the state-invariant admissible action set $U = U(\s), \forall \s \in \Phi$. Theorem~\ref{thm:validation} remains valid for the $\epsilon$-almost controlled safe set validation with a minor difference on how each run of a scenario is propagated. The corresponding safety quantification algorithm also extends the exploration from the given OSS to the combination of the OSS and the testing action space. Definition~\ref{def:epsilon-almost-ssu} can also be extended to the $\epsilon\delta$-almost controlled safe set over a certain $\delta$-covering set of $U(\s)$ similar to the extension from Definition~\ref{def:epsilon-almost-ss} to Definition~\ref{def:epsilondelta-almost-ss}.

In Fig.~\ref{fig:example}, we present a conceptual example illustrating some of the introduced definitions throughout this section. Consider the testing system example shown in Fig.~\ref{fig:example}(a). One considers the SR as an automated emergency brake (AEB) system~\cite{forkenbrock2015nhtsa} equipped vehicle (in gray) and the testing system takes the longitudinal control (i.e. acceleration) of the lead vehicle (in red) as the testing action. The subscripts, 0 and 1, denote the SR follower and the lead vehicle respectively. As a result, the lead vehicle's control policy $\pi_1$ is essentially the testing policy $\pi$ for this testing system. The OSS $O$ (in blue) admits a three-dimensional design including the distance headway (dhw) between the two vehicles and the longitudinal velocities of the two vehicles. Note the dhw is upper bounded by a finite value as a sufficiently large dhw contributes very little value to the safety performance analysis in this example and is also beyond the perception capability of the real-world testing hardware. Other limits are set by the physical capabilities of the vehicles. Other dynamic states (e.g. the acceleration of the SR) and environmental conditions (e.g. road surface friction) are examples of disturbances and uncertainties $w$ in \eqref{eq:ctrl-sys}. A run of a scenario, $\mathcal{R}_{\sigma}(\s_0, k)$ or simply $\mathcal{R}(\s_0)$, is thus a sequence of points in the blue OSS $O$. The set of failure events $\mathcal{C}$ is shown in red as the dhw is negative which further implies a rear-end collision between the two vehicles. In Fig.~\ref{fig:example}(b), two testing algorithms covering the same set of testing actions with different sampling distributions are of Type-I difference by Definition~\ref{def:diff}. In Fig.~\ref{fig:example}(c), two testing algorithms covering different sets of testing actions are of Type-II difference. Two testing algorithms with different feedback policies are also of Type-II difference as shown in Fig.~\ref{fig:example}(d). Note that the upper one in (d) is the exact testing policy of AEB system adopted by the EURO-NCAP testing standard~\cite{van2017euro} (i.e. constant braking-to-stop at $-6$ m/$s^2$). The bottom figure in (d) indicates a neural network-based testing policy. In Fig.~\ref{fig:example}(e), a conceptual example of $\Phi \subset O$ (in green) is illustrated. In particular, $\Phi$ could be (i) almost safe by Definition~\ref{def:epsilon-almost-ss} if one composes one of the feedback testing policies in (d) with \eqref{eq:ctrl-sys}, or (ii) almost \emph{controlled} safe w.r.t. one of the sets of testing actions in (c) by Definition~\ref{def:epsilon-almost-ssu}. Note that in both cases, the SR (the gray subject vehicle) is of higher-risk if the dhw is small and $v_0 \gg v_1$, which explains the difference between $\Phi$ (the green set in (e)) and the OSS $O$ (the blue set in (a)). One can refer to~\cite{weng2021towards,weng2022formal} for more detailed safety analyses regarding this simplified example.

\section{The Comparability of Scenario-based Safety Testing Algorithms}\label{sec:comparability}
The main body of this work starts from a formal analysis of the comparability of various scenario-based safety testing algorithms. The main discovery can be summarized as (i) all Type-I different scenario-based safety testing algorithms perform equally well in the black-box testing configuration, and (ii) the comparability is possible for the non-black-box configuration or the algorithms of Type-II difference. Details are presented as follows. 

\subsection{The Impossibility Theorem for Scenario-based Safety Testing}
Let's start with a deterministic setup with $W = \emptyset$ for~\eqref{eq:ctrl-sys}. The omitted randomness brings extra notation complexity to the analysis without significant impact to the theoretical results.

Recall the termination condition for algorithm $\mathcal{TE}^u$, for two algorithms to obtain the same sequence of costs $g_m$ such that $\mathcal{T}(g_m)=\texttt{True}$, the more adversarial one should achieve the sequence at a higher probability in expectation over the class of all testing systems. Let $\mathbb{P}(g_m \mid f_s, m, \mathcal{TE}^u)$ denote the probability of obtaining a particular aforementioned sequential costs by iterating $m$ runs of scenarios with $\mathcal{TE}^u$ against $f_s$, we have the following impossibility theorem for safety testing.
\begin{theorem}\label{thm:nfl}
Consider an arbitrary pair of scenario-based safety testing algorithms in the form of Algorithm~\ref{alg:te_u}, $\mathcal{TE}^u_1$ and $\mathcal{TE}^u_2$, of the Type-I difference by Definition~\ref{def:diff},
\begin{equation}\label{eq:nfl}
    \sum_{f_s\!\in\!F_s}\!\mathbb{P}(g_m \mid f_s,\!m,\!\mathcal{TE}^u_1) = \sum_{f_s\!\in\!F_s}\!\mathbb{P}(g_m \mid f_s,\!m,\!\mathcal{TE}^u_2).
\end{equation}
\end{theorem}
One can refer to the Appendix~\ref{apx:nfl} for the complete proof of the above theorem. 

Intuitively, Theorem~\ref{thm:nfl} suggests that under an absolute black-box testing configuration by Remark~\ref{rmk:bbox}, \emph{all scenarios-based testing algorithms of Type-I difference perform equally well}. Moreover, revealed by the proof in the Appendix~\ref{apx:nfl}, the expected probability to obtain a particular run of a scenario over all testing systems is also irrelevant of the choice of the testing algorithm $\mathcal{TE}^u$. That is, if a certain state is shown particularly critical for a certain $f_s \in F_s$, there must exist another testing system within $F_s$ against which the state is not as dangerous. Therefore, the explicit exploration sequence of state-actions does not matter and all such algorithms share the same performance with uniform random walk. Note that the black-box configuration by Remark~\ref{rmk:bbox} is the essential key that contributes to the conclusion by Theorem~\ref{thm:nfl} as the uniformly distributed $f_s \in F_s$ ensures that over the space of all testing systems, all states in $O$ are reachable from any state in $O$ given a certain testing action in $U$ and a certain $f_s \in F_s$. Hence there does not exist a particularly critical state or adversarial action as the criticality is dependent upon $f_s$.

The exact empirical proof of Theorem~\ref{thm:nfl} is difficult to establish in practice as the absolute class of black-box testing systems is still rare to encounter. For the similar cause, existing examples with empirical success of various adaptive and importance sampling based testing algorithms are not counter-examples for Theorem~\ref{thm:nfl} either, as the testing system is never a black-box and typically follows a specific function design. 

However, as more complex robots are developed and deployed in complicated working domains, the fundamental problem formulation aligns closer with the underlying conditions that dictate Theorem~\ref{thm:nfl} for safety testing, the impact of Theorem~\ref{thm:nfl}, especially its implications to non-black-box but sufficiently complex systems, has thus been subtly revealed in the field. 

For example, the traditional testing scenarios adopted for safety evaluation of various types of AVs has always been following the same procedure involving (i) studying the real-world vehicle collision events from police reports and data logs, (ii) accumulating experience and expert-knowledge, (iii) directing the built-up expertise to design or re-construct a testing scenario library that prioritizes the exploration of high-risk states and actions subject to vehicle dynamic limits, traffic rules and other constraints. The expert-knowledge built up from step (ii) has also been recently replaced with models and other unsupervised-learning techniques~\cite{wang2018extracting,hauer2020clustering}. The above procedure has been working well for decades against simple AV functionalities such as lane keeping and braking for rear-end collision avoidance~\cite{najm2007pre, forkenbrock2015nhtsa} and is also making its way towards testing more complex AV functionalities~\cite{scanlon2021waymo}. However, problems of directly adapting statistically high-risk scenarios learned from human drivers to an AV of high-level intelligence has also been empirically discussed in the recent literature~\cite{capito2020modeled, hauer2020re}, including one titled ``re-using concrete testing scenario generally is a bad idea"~\cite{hauer2020re}. It indeed is, as suggested by Theorem~\ref{thm:nfl}, a set of high-risk scenarios for one SR does not necessarily remain adversarial against another SR, even if the indicated testing systems belong to the same class of functions. Note that the above setup also fundamentally admits a \emph{regression test} methodology widely adopted in software testing~\cite{leung1989insights,yoo2012regression} and many other industrial applications~\cite{onoma1998regression}. The observed problem with AV applications generalizes to other domains as well.

Finally, for the non-empty $W_s$ to involve in the black-box testing, due to the black-box nature, one simply assumes a uniform distribution of all possible distributions over $W_s$ and the expected outcome involving $w_s$ over each possible distribution. Thanks to the expected probability formulation in~\eqref{eq:nfl}, Theorem~\ref{thm:nfl} should generalize to the stochastic system with unknown disturbances and uncertainties by replacing the specific $g_m$ with the expected $g_m$ over the aforementioned distributions.

\subsection{Towards the Possible Comparability}
To make Theorem~\ref{thm:nfl} invalid, one fundamental requirement is to break the black-box setting. In practice, this is reflected as analyzing only a finite sub-set of testing systems from the complete function space $F_s$. This typically requires some model insights and expert-knowledge to establish the biased prior. Whether the introduced prior is reasonable is problem-specific and occasionally beyond the scope of the engineering discipline. Assume one indeed confines the study to a non-black-box configuration, for a certain $g_m$, the equality of~\eqref{eq:nfl} can be invalid between the pair of algorithms of Type-I difference. If that particular $g_m$ also satisfies $\mathcal{T}(g_m)=\texttt{True}$, not only the pair of algorithms are comparable, one is also better than the other. This is essentially how existing adaptive sampling and learning based adversarial testing methods in the literature~\cite{li2020av,feng2021intelligent} are demonstrated empirically. For example, in~\cite{li2020av}, the authors empirically argue that the proposal being better than the adaptive stress test (AST)~\cite{corso2019adaptive} by showing one forces the vehicle equipped with the Apollo self-driving stack to a collision with less number of sampled runs of scenarios than AST. The selection of a specific software stack, Apollo, makes the evaluation non-black-box. 

Another option to make Theorem~\ref{thm:nfl} invalid is to extend the difference through the selection of the testing action set $U$ or the testing policy $\pi$ (i.e. switching from the Type-I difference to Type-II difference by Definition~\ref{def:diff}). Note the inclusion of different $U$ or $\pi$ leads to various biased coverage of proper subsets of $G$. One can no longer rely on the notion of the expected probability of reaching the same $g_m$ to help claim one algorithm being different from the other as the acquirable sets of costs for different algorithms are different. Moreover, the fact that two algorithms are different does not necessarily indicate that one is better than the other. The formal notion of the relatively better testing algorithm as well as the constructive solution to justify the performance discrepancies among various testing approaches are further addressed in the next section.

\section{Towards a Provably More Aggressive Scenario-based Safety Testing Algorithm }\label{sec:optimality}
In this section, we focus on safety testing algorithms of Type-II difference. That is, the only difference between different algorithms, by Definition~\ref{def:diff}, is the adopted testing action set and policy. As we have clarified in the last section, although the algorithms perform differently, one can no longer rely on the notion of acquisition probability towards the same output to justify the optimality comparison.  

Recall the discussion from Section~\ref{sec:preliminaries}, the safety metric is not only a performance characterization of the set of subject policies or the testing system, but also a justification of the testing action set and testing policy adopted by the algorithms. For example, the SR tested relatively safe in the system $f_s$ against $\pi$ is equivalent of saying the testing policy $\pi$ is not sufficiently aggressive to force the SR to exhibit unsafe behaviors or $f_s$ to converge to risky outcomes.

The above idea has been extensively used with various safety metrics justifying the testing methods' aggressiveness such as the magnitude of jerk values~\cite{bellem2018comfort} and the observed risk (failure rate)~\cite{li2020av}. Intuitively, if a certain safety evaluation algorithm $\mathcal{TE}_1$ is deemed more aggressive than another algorithm $\mathcal{TE}_2$ (assuming they are of Type-II difference), one expects the SR tested safe by $\mathcal{TE}_1$ to also be safe with $\mathcal{TE}_2$. This fundamentally admits a \emph{transitive property}. However, the aforementioned metrics, as we should address theoretically later in this section and empirically in Section~\ref{sec:case}, can occasionally fail to satisfy the transitive property. 

\subsection{Define Aggressiveness}
One unique property of testing algorithms of Type-II difference is that they cover different subsets of states and actions. It is thus a natural idea to use that subset, or a certain unique part of that subset, to characterize the performance of a testing algorithm. This leads to the proposal of adopting a safe set perspective to help defining the aggressiveness comparison between safety evaluation algorithms of the Type-II difference, which can be formally addressed by the following definitions.
\begin{definition}\label{def:agg_pu}
    Let $\Phi_i \subset O, i\in\{1,2\}$. $\Phi_i$ is the $\epsilon$-almost (controlled) safe set obtained from testing a certain SR in $f_s$ uniformly distributed in a set of testing systems $\bar{F_s}\subseteq F_s$ against the testing policy $\pi^i$ (or the testing action set $U^i$). $\pi^1$ (or $U^1$) is deemed $\epsilon$-almost (controlled) more aggressive than $\pi^2$ (or $U^2$) if
    \begin{equation}\label{eq:agg}
        \Phi_1 \subset \Phi_2.        
    \end{equation}
    Similarly, $\pi^1$ (or $U^1$) is $\epsilon$-almost (controlled) less aggressive than and equally aggressive with $\pi^2$ (or $U^2$) if $\Phi_1 \supset \Phi_2$ and $\Phi_1 = \Phi_2$, respectively. 
\end{definition}
\begin{definition}\label{def:agg}
    Let $\mathcal{TE}_1$ and $\mathcal{TE}_2$ be of Type-II difference by Definition~\ref{def:diff} (i.e. $\pi^1 \neq \pi^2$ or $U^1 \neq U^2$). The testing policy $\mathcal{TE}_1$ is deemed $\epsilon$-almost (controlled) more aggressive than $\mathcal{TE}_2$ if $\pi^1$ (or $U^1$) is deemed $\epsilon$-almost (controlled) more aggressive than $\pi^2$ (or $U^2$) by Definition~\ref{def:agg_pu}.
\end{definition}
Note that the aggressiveness comparability defined above is w.r.t. the different components between the two algorithms (i.e. different $U$, or different $\pi$). 

It is immediate that the commonly used risk based justification forms a necessary but insufficient condition for the aggressiveness justification by Definition~\ref{def:agg}. That is, a more aggressive testing policy by Definition~\ref{def:agg} implies lower-risk, but the lower-risk policy does not necessarily indicate the proper subset relation required by Definition~\ref{def:agg} as it could also be the case where the two almost-safe sets are different (i.e. $\Phi_1 \setminus \Phi_2 \neq \emptyset$ and $\Phi_2 \setminus \Phi_1 \neq \emptyset$). Moreover, the intuitive equivalence between large action magnitude and high aggressiveness is also problematic, as the cardinality of the almost safe set is not necessarily monotonically related to the magnitudes of control actions. We will see more examples for both of the above deficiencies later in Section~\ref{sec:case}.

Moreover, we argue that the $\epsilon$-almost safe set justifies the testing algorithm's performance in a complete way that better aligns with the intuitive expectation of aggressiveness. It is \emph{complete} in the sense it casts the performance of a certain safety testing algorithm over a multi-dimensional feature space characterizing various set properties such as the cardinality, the coverage, and the perimeter shape. The obtained outcome specifies not only the risk (through $\epsilon$) but also where in the testing state domain the algorithm exhibits the risk in a provably unbiased manner. $\mathcal{TE}_1$ is thus different from $\mathcal{TE}_2$ in various ways related to the mentioned set features. It is also quite \emph{intuitive} as the basic transitive property is guaranteed by definition given the specific comparable aggressiveness only occurs under the strict condition of~\eqref{eq:agg}. That is, if the SR is tested safe at every $\s$ in $\Phi$ against $\mathcal{TE}_1$ deemed almost more aggressive, within a controllable level of accuracy (through $\epsilon, \beta$, and $\delta$ as we should mention later), it is also safe at every $\s$ in $\Phi$ against $\mathcal{TE}_2$.

Taking advantage of the proposed Definition~\ref{def:agg}, some concrete solutions that help making the formal aggressiveness comparison among algorithms are discussed next.

\subsection{Formal Aggressiveness Comparison}
The most intuitive approach to make the aggressiveness comparison between two safety evaluation algorithms, by Definition~\ref{def:agg}, is through deriving the $\epsilon$-almost safe sets for both of them with the same $\epsilon$ and other hyper-parameters. However, the advantage one considers an adversarial alternative in the first place is to improve the testing efficiency. The accurate derivation of the $\epsilon$-almost safe sets for both algorithms fails that purpose. Moreover, an algorithm might be re-directed to various other testing purposes and to work with different cost function designs and termination conditions. The computational effort to justify the aggressiveness should be kept at a practically acceptable level.

One immediate alternative is to adopt one of the resolution-complete algorithms such as the ones incorporating the $\delta$-covering set discussed in Section~\ref{sec:preliminaries} and other literature~\cite{weng2021formal}. The improved computational efficiency is achieved through the trade-off with the set accuracy in terms of the resolution and probabilistic completeness.

However, even with the resolution relaxation, one still has to fully quantify the $\epsilon\delta$-almost (controlled) safe sets for both algorithms to justify the relative aggressiveness. In the remainder of this section, another formal aggressiveness justification algorithm is proposed, which further improves the above alternative. 

Intuitively, suppose one has already quantified a certain $\epsilon$-almost safe set, $\Phi_1$, for $\mathcal{TE}^{\pi}_1$ which is one of the two safety evaluation algorithms to be compared of Type-II difference with confidence level at least $1-\beta$. Consider the claim of ``$\mathcal{TE}^{\pi}_1$ is $\epsilon$-almost more aggressive than or equally aggressive with $\mathcal{TE}^{\pi}_2$" and $\frac{\ln{\beta}}{\ln{(1-\epsilon)}}$ sampled runs of scenarios initialized from $\Phi_1$ but tested against $\mathcal{TE}^{\pi}_2$. If all sampled state trajectories stay in $\Phi_1$, it is immediate that the claim is true by Definition~\ref{def:agg}. If any run of a scenario ends up in $\mathcal{C}$, then the claim is false by Definition~\ref{def:epsilondelta-almost-ss}. One last possible outcome is that some sampled runs of scenarios initialized in $\Phi_1$ reach outside $\Phi_1$ but are not reaching $\mathcal{C}$. It is in general difficult to justify the property of the outreach part of states without sampling runs of scenarios initialized from those states. However, if the outreach part of states are only of low probability of occurrence, one might not need to test those states anyway. This inspires the following theorem that sheds light on an approach that requires only a limited added runs of scenarios to validate the aggressiveness comparison. The theorem is presented w.r.t. $\mathcal{TE}^{\pi}$ and the generalization to $\mathcal{TE}^{u}$ is straight-forward.
\begin{theorem}\label{thm:agg}
Given an OSS $O \subset S$ and an $\epsilon$-almost safe ODD $\Phi_1 \subset O$ with confidence level $\beta \in (0,1]$ and $\epsilon \in (0,1]$ by Definition~\ref{def:epsilon-almost-ss} obtained with $f_s$ uniformly distributed in $\bar{F}_s \subset F_s$ and $\pi_1$ from $\mathcal{TE}^{\pi}_1$. Let $\pi_2$ from $\mathcal{TE}^{\pi}_2$ be a different testing policy from $\pi_1$. $\mathcal{TE}^{\pi}_1$ and $\mathcal{TE}^{\pi}_2$ are of Type-II difference by Definition~\ref{def:diff}. Let $p_S(\s)$ be the probability mass function of $\s$ in $S$. Consider consecutively $N$ runs of scenarios obtained from testing the SR in the uniformly distributed systems in $\bar{F}_s$ using $\pi_2$ with the state initialization being i.i.d. w.r.t. the underlying distribution and $\{\s_i\}_{i\in\Z_N} \subset \Phi_1$. Let $\Phi'_1 = \Phi \cup \{\mathcal{R}_{\sigma}(\s_i)\}_{i\in\Z_N}$. With confidence level at least $1-\beta$, the nominal testing policy $\mathcal{TE}^{\pi}_1$ is $\epsilon$-almost more aggressive than or equally aggressive with $\mathcal{TE}^{\pi}_2$ by Definition~\ref{def:agg} if
\begin{subequations}\label{eq:agg_val}
    \begin{align}
        & N \geq \frac{\ln{\beta}}{\ln{(1-\epsilon)}}, \label{eq:agg_val1} \\
        & \Phi_1 \subseteq \Phi'_1 \cap \mathcal{C} = \emptyset, \label{eq:agg_val2} \\
        & \frac{\sum_{\s \in \Phi'_1 \setminus \Phi_1} p_S(\s)}{\sum_{\s \in \Phi'_1} p_S(\s)} < \epsilon. \label{eq:agg_val3}
    \end{align}
\end{subequations}
\end{theorem}
One can refer to Appendix~\ref{apx:agg} for the proof of the above theorem. Note that theorem~\ref{thm:agg} can be combined with the $\epsilon\delta$-almost safe set to form a provably unbiased and computationally efficient aggressiveness comparison algorithm justifying if $\mathcal{TE}^{\pi}_1$ is more aggressive than or equally aggressive with $\mathcal{TE}^{\pi}_2$ at an acceptable level of resolution accuracy dictated by the defined $\delta$. The details are presented in Algorithm~\ref{alg:agg_val} and can be easily generalized to the case with $\mathcal{TE}^{u}$. Let $\mathcal{QNT}_{\mathcal{TE}^{\pi}}(O, \bar{F}_s, \epsilon, \beta, \delta)$ be one of the $\epsilon\delta$-almost safe set quantification algorithms~\cite{weng2021towards,weng2021formal,weng2022formal,sdq_tools} taking the testing policy from $\mathcal{TE}^{\pi}$ against the SR in $f_s$ uniformly distributed in a certain $\bar{F}_s$ (one can refer to Appendix~\ref{apx:qnt} for more details).

\begin{algorithm}[H]
    \begin{algorithmic}[1]
    \State {\bf Given:} $\epsilon \in (0,1]$, $\beta\in(0,1]$, $\delta \in \R_{\geq0}^n, O\subset S$, $\mathcal{C}$, two different safety evaluation algorithms $\mathcal{TE}^{\pi}_1$ and $\mathcal{TE}^{\pi}_2$ admitting the Type-II difference
    \State {$(\Phi_{\delta}^1, \Phi_s^1) = \mathcal{QNT}_{\mathcal{TE}^{\pi}_1}(O, \bar{F}_s, \epsilon, \beta, \delta)$ }
    \State {\bf Initialize: } \texttt{AGG}=\texttt{False}, $N = 0, \Phi_{\delta}^2=\Phi_{\delta}^1, \Phi_s^2=\Phi_s^1$
    \State{{\bf While} $N<\frac{\ln{\beta}}{\ln{(1-\epsilon)}}$:}
    \State{\ \ \ \ Get $\s_0 \in \Phi_s^1$, collect $R = \mathcal{R}(\s_0)$ with $\mathcal{TE}^{\pi}_2$}
    \State{\ \ \ \ {\bf If} $R\cap \mathcal{C} \neq \emptyset$}
    \State{\ \ \ \ \ \ \ \ {\bf break}}
    \State{\ \ \ \ {\bf If} $R \setminus \Phi_{\delta}^2\neq \emptyset$}
    \State{\ \ \ \ \ \ \ \ {\bf For} $\s'$ in $R'= R \setminus \Phi_{\delta}^2$ {\bf do}}
    \State{\ \ \ \ \ \ \ \ \ \ \ \ {\bf If} $\s' \notin \Phi_{\delta}^2$}
    \State{\ \ \ \ \ \ \ \ \ \ \ \ \ \ \ \ $\Phi_{\delta}^2=\Phi_{\delta}^2 \cup \mathcal{N}_{\delta}(\s')$, $\Phi_{s}^2=\Phi_{s}^2 \cup \{\s'\}$}
    \State{{\bf If} $\frac{\sum_{\s \in \Phi_s^2 \setminus \Phi_s^1} p_S(\s)}{\sum_{\s \in \Phi_s^2} p_S(\s)} < \epsilon$}
    \State{\ \ \ \ \texttt{AGG}=\texttt{True}}
    \State{{\bf Else}}
    \State{\ \ \ \ $(\Phi_{\delta}^2, \Phi_s^2) = \mathcal{QNT}_{\mathcal{TE}^{\pi}_2}(O, \bar{F}_s, \epsilon, \beta, \delta)$}
    \State{\ \ \ \ {\bf If} $\Phi_{\delta}^2 \supseteq \Phi_{\delta}^1$}
    \State{\ \ \ \ \ \ \ \ \texttt{AGG}=\texttt{True}}
    \State {{\bf Output:} \texttt{AGG}}
    \end{algorithmic}
    \caption{Comparing aggressiveness between $\mathcal{TE}^{\pi}_1$ \& $\mathcal{TE}^{\pi}_2$} \label{alg:agg_val}
\end{algorithm}

Overall, Algorithm~\ref{alg:agg_val} starts from characterizing the $\epsilon\delta$-almost safe set using the testing policy from one of the algorithms to be compared (line 2). W.l.o.g, let $\mathcal{TE}^{\pi}_1$ be the presumably more aggressive one. We have the obtained $\delta$-covering set $\Phi_{\delta}^1$ and the centroids $\Phi_s^1$. Next, one starts the collection of at least $\frac{\ln{\beta}}{\ln{(1-\epsilon)}}$ sampled runs of scenarios with the initialization state being i.i.d. w.r.t. the underlying distribution of $\Phi_s^1$ obtained from the known distribution over $O$ ($\Phi_s^1 \subset O$). The above step should end up with one of four outcomes. (i) If one encounters a failure run of a scenario initialized from $\Phi_s^1$ following the testing algorithm $\mathcal{TE}_2$ at the $N$-th trial and $N < \frac{\ln{\beta}}{\ln{(1-\epsilon)}}$ (line 6), the SR is not sufficiently safe in $\Phi_{\delta}^1$ against $\mathcal{TE}_2$, neither is $\mathcal{TE}_1$ almost more aggressive than or equally aggressive with $\mathcal{TE}_2$ (as $\mathcal{TE}_2$ captures a failure event that $\mathcal{TE}_1$ fails to capture within the same level of effort). On the other hand, if the SR successfully ``survives" $\frac{\ln{\beta}}{\ln{(1-\epsilon)}}$ runs of scenarios against $\mathcal{TE}_2$, we have two other possible outcomes. (ii) If the union of the runs of scenarios remain inside $\Phi_{\delta}^1$ (i.e. line 8-11 were never executed and $\Phi_{\delta}^1=\Phi_{\delta}^2$), $\mathcal{TE}_1$ is $\epsilon$-almost more aggressive than $\mathcal{TE}_2$ by Definition~\ref{def:agg} within the resolution accuracy level justified by $\delta$ (i.e. \texttt{AGG}=\texttt{True}). (iii) Otherwise, the above aggressiveness comparison still remains valid if $\Phi_{\delta}^2$ does not deviate too much from $\Phi_{\delta}^1$ with the allowed deviation formally justified by~\eqref{eq:agg_val3} using the sets of centroids $\Phi_s^1$ and $\Phi_s^2$ (line 12-13). (iv) The last possible outcome is that $\Phi_{\delta}^1$ differs from $\Phi_{\delta}^2$ significantly (line 15-17), one thus relies on the formal quantification (see Algorithm~\ref{alg:qnt} in Appendix~\ref{apx:qnt}) of the $\epsilon\delta$-almost safe set for $\mathcal{TE}^{\pi}_2$ to give the comparison outcome.

This section is concluded with a remark emphasizing that Algorithm~\ref{alg:agg_val}, similar to the other safe set quantification variants~\cite{weng2021towards}, theoretically suffers from the Curse-of-Dimensionality (CoD). Relaxing the resolution and probabilistic accuracy through smaller $\delta$, $\epsilon$, and $\beta$ is a feasible fix of relieving the impact of CoD in practice. This is also the method adopted throughout Section~\ref{sec:case}. One other practically efficient method is to initialize Algorithm~\ref{alg:agg_val} with the OSS $O$ being closer to the potential almost safe set. Such non-black-box insights are typically achieved through (i) expert knowledge and (ii) model-specific understandings of the system. Developing a more fundamentally effective solution for the CoD challenge is of future interest and it is out of the scope of this paper.

\section{Case Studies}\label{sec:case}
The safety evaluation of two types of robots of different mechanical nature and intended functionalities are presented in this section. For the bipedal robot, Cassie, one studies the safety property of stable locomotion control (not falling over) against external disturbances. For the decision-making algorithms of vehicles, one evaluates the collision avoidance capability against other testing vehicles following different forms of testing strategies.

We emphasize that the studied cases are not only to empirically demonstrate a particular algorithm (as Algorithm~\ref{alg:agg_val} is only one of the several contributions presented in the previous two sections). Also, as addressed in Section~\ref{sec:comparability}, the empirical proof of Theorem~\ref{thm:nfl} is difficult to have. However, its implications will still be observed from both cases. 

\subsection{Case Study with Bipedal Locomotion Controllers of Cassie}\label{sec:case_cassie}

\begin{figure}[t]
    \centering
    \includegraphics[width=0.49\textwidth]{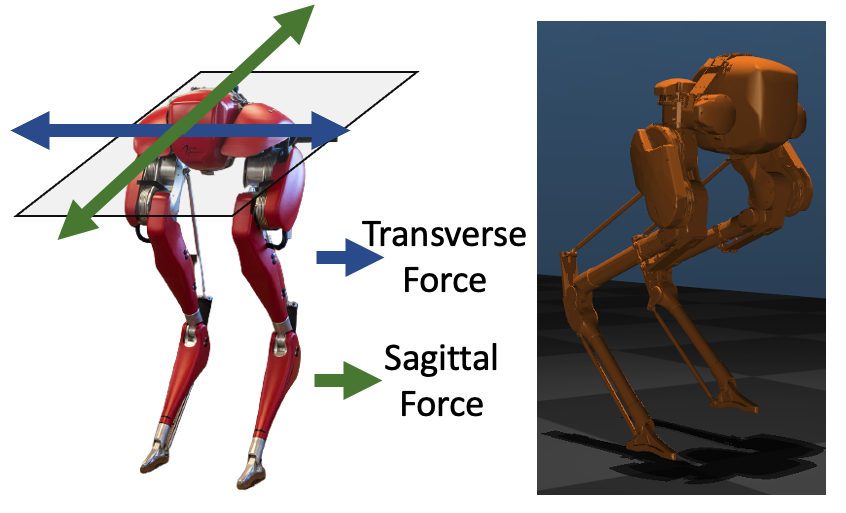}
    \caption{The Cassie robot in real-world (left) and MuJoCo simulator (right) with annotations of the transverse and sagittal domains studied by this paper.}
    \label{fig:cassie}
    \vspace{-5mm}
\end{figure}

The first case focuses on the class of dynamic locomotion controllers of a 20-degree-of-freedom bipedal robot named Cassie (see Fig.~\ref{fig:cassie}). Consider the class of locomotion controllers that seeks to track a constant desired forward-walking velocity of $0.4$ m/s after the scenario run initialization. The robot is expected to converge to the desired velocity from any instantaneous velocity and any acceleration mode with stable walking gaits without falling over (i.e. the physical contact between the ground and any part of the SR other than the two feet) against external forces applied to the center of gravity (CG) of the robot's torso. This leads to a two-dimensional OSS design as $S = V_c \times Q_c$ with $V_c=[-0.2,0.8] \subset \R$ and $Q_c=\{\texttt{deceleration}, \texttt{steady-state}, \texttt{acceleration}\}$ which denote the set of step velocities and the set of step acceleration modes, respectively. Due to the mechanical nature of Cassie, it is infeasible for the robot to precisely maintain a certain walking speed. As a result, we define the step velocity and acceleration as the filtered average velocity and acceleration, which is a common setup in the field~\cite{castillo2020hybrid,Siekmann-RSS-20,weng2022leeged}. The three discrete modes in $Q_c$ are further defined based on the signs of the step acceleration value. Note that the OSS design considers important state features that are crucial to the desired function of interest. Some other states such as the transverse velocity and the specific acceleration value are considered as part of the disturbances and uncertainties assumed to follow a certain unknown distribution. Moreover, w.l.o.g, we assume the state initialization uniformly distributes in the defined OSS. One can replace this assumption with other types of distributions if applicable.

Consider a periodic push-over testing policy of applying a certain external force to the CG of the robot's torso at 0.5 Hz. The external force admits the testing action set of $F_x \times F_y$ with the aforementioned external forces applied along the sagittal domain ($F_x=\{-40,-39,\dots,40\}$ N) and the transverse domain ($F_y=\{-20,-19,\dots,20\}$ N). 

Each run of a scenario first initializes the robot at the given velocity $v_0 \in V_c$ (within a tolerance of $\pm 0.1$ m/s) and the given acceleration mode (by heuristically configuring a sequence of desired velocities to track). The test then proceeds with periodically applying and relieving the selected external force $f_e \in F_x \times F_y$ with equal time duration (1 second) at 0.5 Hz. We also have the maximum time duration of a single scenario run as $10$-second and the data acquisition frequency is 10 Hz (i.e. $k=100$). The above testing procedure has been discussed in~\cite{weng2022leeged} w.r.t. a similar application. Let falling-over be the only failure event of concern which defines $\mathcal{C}$. All tests are performed in the MuJoCo simulator~\cite{todorov2012mujoco}, using the same environment configuration shared by the developers of the SR's locomotion controllers studied by this paper as we will introduce later. 

For the above mentioned desired locomotion function, we adopt two different locomotion controllers for the SR including one learned through recurrent neural network and proximal policy gradient (RPPO)~\cite{Siekmann-RSS-20} and another policy derived using hybrid zero dynamics inspired evolution strategies (HZDES)~\cite{castillo2020hybrid,castillo2022reinforcement}. Both policies have shown empirically competitive performance in the intended functionality of stable velocity tracking without falling-over. One can refer to~\cite{weng2022leeged} for the safety performance comparison between the two controllers with various OSS designs.

Note that both Type-I different and Type-II different testing algorithms are implemented in this study. The Type-I different algorithms have the same admissible set of testing forces $F_x \times F_y$ as mentioned above. If the push-over testing policy is parameterized on the selected force and the force is set without state-dependency before each run of a scenario, one can consider the testing policy as part of the testing system $f_s$. The studied algorithms are thus fundamentally of Type-I difference. However, the black-box configuration no longer applies with the push-over testing policy involved, yet the implications of Theorem~\ref{thm:nfl} are still available. Moreover, if the two algorithms have different set of testing forces, they become Type-II different algorithms by Definition~\ref{def:diff}.  

\begin{figure}
    \centering
    \includegraphics[width=0.5\textwidth]{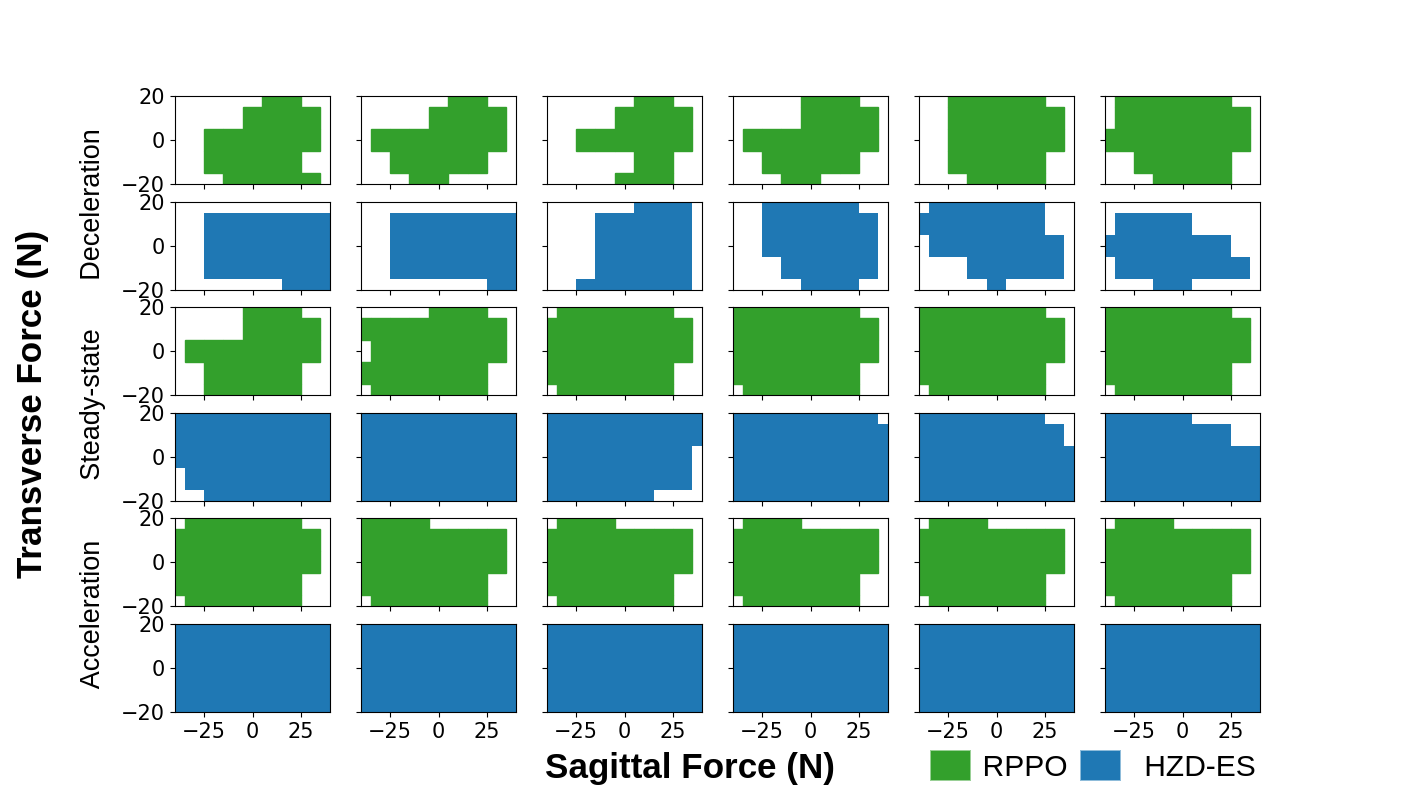}
    \caption{Some sub-space slices of the $\epsilon\delta$-almost controlled safe set obtained for both RPPO and HZDES. Each sub-figure is a $\delta_u$-covering set ($\delta_u=[10,10]$) of the state-dependent admissible testing action set $U(\s) = F_x(\s) \times F_y(\s)$ on which the set of states ($V_c \times Q_c$) forms an $\epsilon\delta_s$-almost controlled safe set ($\epsilon=0.001, \delta_s=[0.1,1]$) with confidence level at least $0.9999$ ($\beta=0.0001$). The $\delta_u$-covering sets for testing RPPO and HZD-ES are shown in green and blue, respectively. Note that the three modes in $Q_c$ are processed as the integer set $\Z_3$, hence the selected $\delta_s[2]=1$ ensures the accurate coverage of all acceleration modes. For all sub-figures, the positive values indicate pushing backward and towards the left of the SR, respectively.}
    \label{fig:cassie_phi}
    \vspace{-5mm}
\end{figure}

\subsubsection{Testing Cassie with Type-I different algorithms}
We start with an experiment that sheds light on Theorem~\ref{thm:nfl}. Fig.~\ref{fig:cassie_phi} shows some of the subspace slices of the quantified $\epsilon\delta$-almost controlled safe sets comparing the RPPO policy against the HZDES. With the same OSS and the same initial testing action space, the two subject policies, RPPO and HZDES, exhibit significant difference in handling various external forces at different states (step velocities and acceleration modes). In particular, RPPO is more vulnerable against external forces applied from the rear-left side of the robot, especially at low forward-walking speeds and walking backwards. That is, consider some safety evaluation algorithms exploring the admissible action set $F_x \times F_y$ at different orders (i.e. Type-I difference) in searching for a certain amount of unsafe states. It is immediate that the one that prioritizes the search in the second quadrant of $F_x \times F_y$ is more efficient against RPPO than with HZDES. On the other hand, the one that prioritizes the search in the first quadrant of $F_x \times F_y$ makes an efficient testing strategy for the HZDES. 

In general, note that RPPO and HZDES are simply two concrete examples from a large function space of unknown locomotion controllers for Cassie, and Cassie is also a specific example from a large space of legged robot dynamics. In the black-box testing configuration where the legged robot could exhibit all possible behaviors with all kinds of mechanical designs (e.g. the bipedal, the quadruped, and the humanoid robot), the best testing sequence of state-actions does not exist, which is proved by Theorem~\ref{thm:nfl}. 

Intuitively, one might expect the algorithm prioritizing the search of high-magnitude forces being relatively efficient w.r.t. a certain falsification purposed termination condition. This is technically incorrect as the legged robot admits the mechanical and controller designs that do not necessarily response linearly w.r.t. the magnitude of the external force (e.g. producing lower stiffness against smaller forces). It is an expected behavior of a ``reasonable" dynamical response, but assuming a particular type of biased behavior is never a prior of any black-box testing algorithm. In fact, as illustrated by the sub-figure for the deceleration mode of RPPO at $0.2$ m/s (first row, third column) in Fig.~\ref{fig:cassie_phi}, given the zero sagittal force, the robot remains safe against the 20 N transverse force but falls over against the 10 N one. As the two selected subject policies are among state-of-the-art, the aforementioned problem is still rare to encounter. However, in the general black-box testing configuration, as suggested by Theorem~\ref{thm:nfl}, safety evaluations algorithms of the Type-I difference does not have a work-for-all adversarial testing strategy.

\begin{figure}
    \centering
    \includegraphics[width=0.5\textwidth]{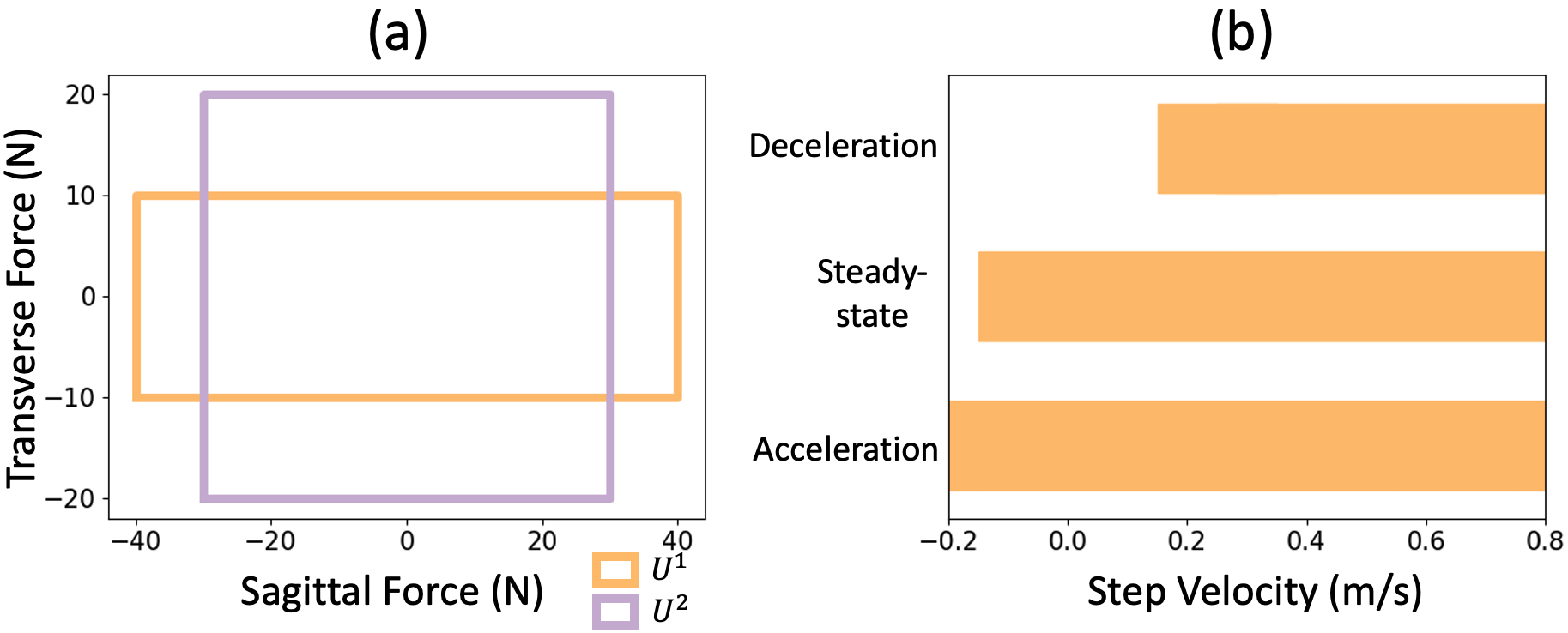}
    \caption{The two adopted state-invariant testing action sets in (a) and the $\epsilon\delta$-almost controlled safe set for $U^1$ against the subject policies, RPPO and HZDES, selected with equal probability at each test run with $\epsilon=0.001, \beta=0.0001, \delta=[0.1,1]$.}
    \label{fig:cassie_uts}
    \vspace{-5mm}
\end{figure}
\begin{table}[b]
    \centering
    \caption{Efficiency (total number of runs of scenarios) comparison between Algorithm~\ref{alg:agg_val} and the direct $\epsilon\delta$-almost safe set quantification against RPPO with $\delta=[0.1,1], \beta=0.0001$. The number of samples for $\mathcal{QNT}_{\mathcal{TE}^{u}_1}(O, \text{RH}, \epsilon, \beta, \delta)$ is not included in the presented values as it is shared by both approaches.}
    \label{tab:efficiency}
    \resizebox{0.49\textwidth}{!}{%
    \begin{tabular}{l|c|c|c}
    \hline
    $\epsilon$ & 0.1 & 0.01 & 0.001 \\\hline
    Algorithm~\ref{alg:agg_val} with $U$  & 88 & 917 & 9206  \\  \hline
    $\mathcal{QNT}_{\mathcal{TE}^{u}_2}(O, \text{RH}, \epsilon, \beta, \delta)$ & 271 & 3061 & 18482 \\ \hline
    \end{tabular}%
    }
\end{table}

\subsubsection{Testing Cassie with Type-II different algorithms}
Moreover, consider the SR Cassie uniformly selecting the locomotion controller from $\{\text{RPPO}, \text{HZDES}\}$ at each run of a scenario. Let it be tested against two safety evaluation algorithms with different $U$s (i.e. Type-II difference), as shown in Fig.~\ref{fig:cassie_uts}(a) denoted as $U^1$ and $U^2$. It is immediate that the control action magnitude cannot be taken as the aggressiveness indicator given each $U$ ``specializes" at a certain sub-region of actions.

The $\epsilon\delta$-almost controlled safe set against the testing action set of $U^1$ is shown in Fig.~\ref{fig:cassie_uts}(b). The corresponding almost controlled safe set of $U^1$ is not calculated directly as one relies on Algorithm~\ref{alg:agg_val} to make the aggressiveness comparison.

The derivation of the illustrated almost safe set in Fig.~\ref{fig:cassie_uts}(b) follows the one presented in the Appendix~\ref{apx:qnt} and is executed at line 2 in Algorithm~\ref{alg:agg_val}. By proceeding with the algorithm of deploying the obtained $\epsilon\delta$-almost safe set $\Phi_1$ to be tested against $U^2$, line 12 is triggered and the testing algorithm with $U^1$ is thus deemed $\epsilon$-almost more aggressive than or equally aggressive with the one using $U^2$ at the resolution accuracy level justified by $\delta$ with confidence level at least $0.9999$. Moreover, the empirical advantage of Algorithm~\ref{alg:agg_val} with $U$ (a direct extension of Algorithm~\ref{alg:agg_val} replacing the testing policy $\pi$ with the testing action space) compared with the direct quantification algorithm for aggressiveness comparison is shown in Table~\ref{tab:efficiency}. Note that $\mathcal{QNT}_{\mathcal{TE}^{u}_2}(O, \text{RH}, \epsilon, \beta, \delta)$ denotes the $\epsilon\delta$-almost controlled safe set quantification algorithm with the testing action set $U^2$ against the SR uniformly selecting controllers from $\text{RH}=\{\text{RPPO}, \text{HZDES}\}$. It is immediate from Table~\ref{tab:efficiency} that the proposed method is more sampling efficient than the direct quantification of the $\epsilon\delta$-almost controlled safe set for the aggressiveness comparison propose.

Note that for this particular example, the risk-aggressiveness equivalence is empirically valid. Consider $1000$ uniformly sampled runs of scenarios, the algorithm with $U^1$ and $U^2$ end up with the observed failure rate of $17\%$ and $10\%$, respectively. However, the intuitive risk-aggressiveness equivalence is not always valid as we have analytically discussed in Section~\ref{sec:optimality} and will empirically illustrate towards the end of the next sub-section.

\begin{figure*}[!ht]
    \centering
    \includegraphics[trim={1cm 0cm 0cm 0cm},clip,width=0.99\textwidth]{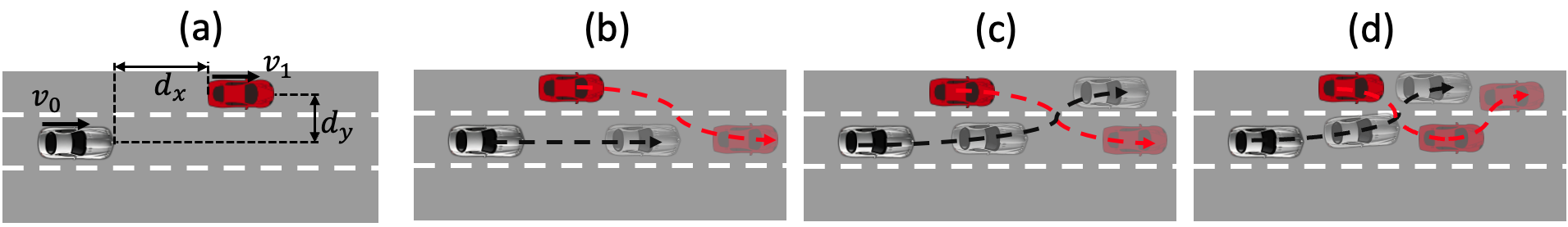}
    \caption{The OSS configuration for the case study with decision-making modules (sub-figure (a)) and some conceptual example runs of scenarios between some of the selected decision-making module and testing policy pairs including (b) the POV performs a lane-change and braking maneuver that forces the SV to brake-to-stop typically observed between $\pi_0^c$ and $\pi_1 \in \{\pi_1^h, \pi_1^p, \pi_1^e\}$, (c) the POV follows a similar maneuver as in (a) but the SV performs a lane-change for collision avoidance and speed maintenance typically observed between $\pi_0^a$ and $\pi_1 \in \{\pi_1^h, \pi_1^p, \pi_1^e\}$, and (d) the POV stays on the intended trajectory of the SV while forcing the $d_x$ to decrease typically observed between $\pi_0^a$ and $\pi_1 \in \{\pi_1^p, \pi_1^e\}$}.
    \label{fig:ads_oss_exp}
    \vspace{-5mm}
\end{figure*}

\begin{figure*}
\centering
\begin{subfigure}{.49\textwidth}
  \centering
  \includegraphics[trim={0cm 0cm 2cm 2cm},clip,width=\linewidth]{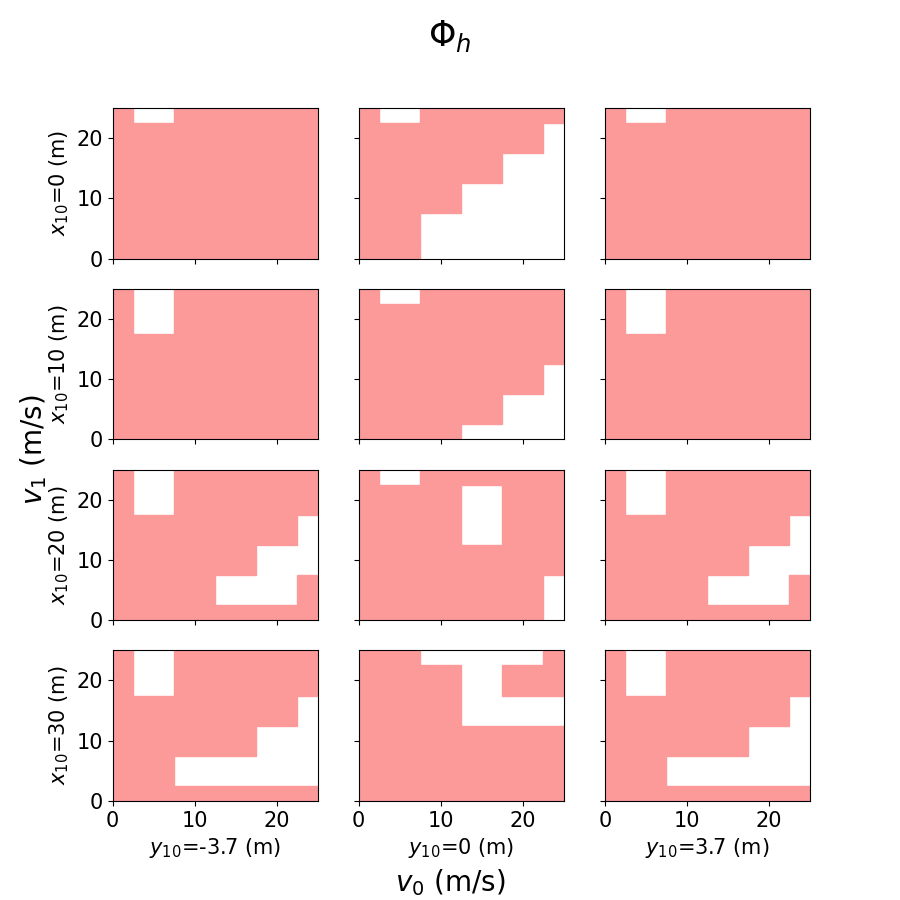}
  \caption{Slices of $\Phi_h^a$.}
  \label{fig:phi_h}
\end{subfigure}%
\hfill
\begin{subfigure}{.49\textwidth}
  \centering
  \includegraphics[trim={0cm 0cm 2cm 2cm},clip,width=\linewidth]{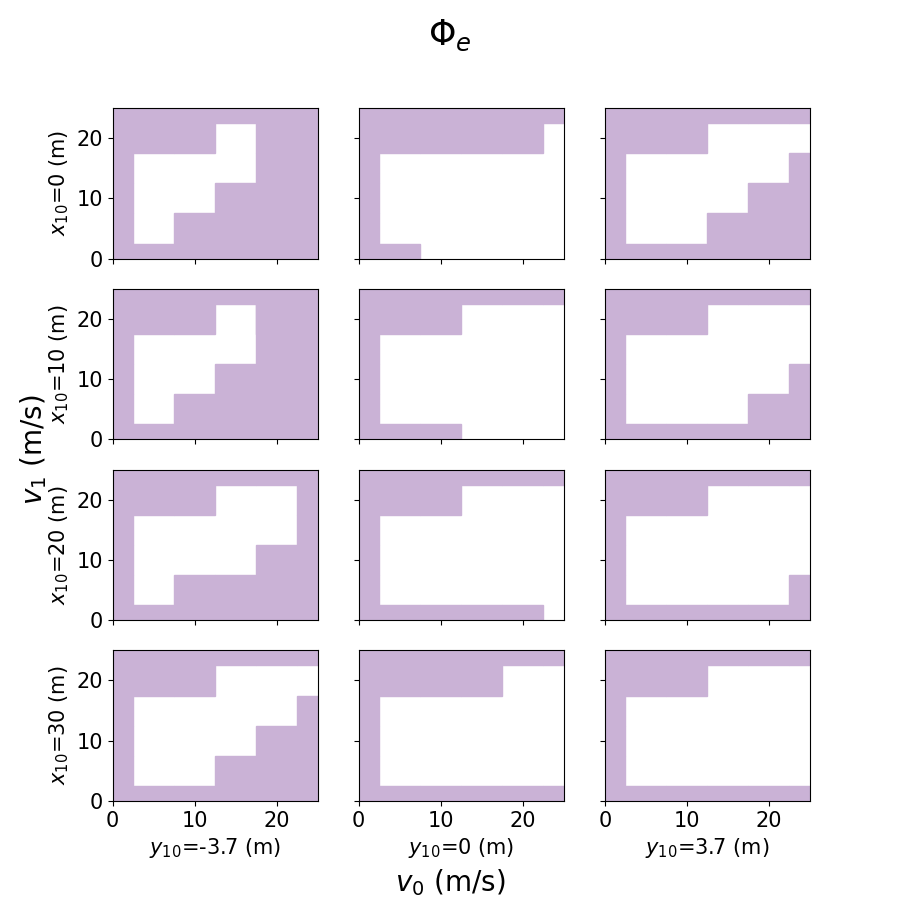}
  \caption{Slices of $\Phi_e^a$.}
  \label{fig:phi_e}
\end{subfigure}
\caption{Some examples of the sub-space slices of the $\epsilon\delta$-almost safe sets (denoted as the colored regions in each all sub-figures) obtained from testing $\pi_0^a$ using the testing policies $\pi_1^h$ and $\pi_1^e$ ($\epsilon=0.01, \beta=0.0001, \delta=[5,3.7,5,5]$). }
\label{fig:ads_phi}
\vspace{-5mm}
\end{figure*}

\subsection{Case Study with Decision-Making Modules}
The second case is concerned with the application of safety evaluation of vehicle decision-making modules. In this study, we consider two decision-making modules, $\pi_0^c$ and $\pi_0^a$, affiliated with the Subject Vehicle (SV). Similar to the example in Fig.~\ref{fig:example}, the subscript $0$ denotes the SV-related property, and the subscript $1$ is used for the other lead vehicle as we shall introduce later. $\pi_0^c$ admits an adaptive-cruise-control (ACC) type driving policy that always tracks the lane-center, adapts its speed to the lead Principal Other Vehicle (POV) constrained by the speed limit, and takes the longitudinal braking as the only collision-avoidance maneuver (i.e. the subject vehicle never changes lanes). The described functionality is achieved through the Intelligent Driving Model (IDM)~\cite{treiber2013traffic} which is one of the most well-adopted longitudinal driving models in the field~\cite{feng2020testing,weng2021towards,capito2020modeled,weng2022formal}. On the other hand, $\pi_0^a$ extends the capability of $\pi_0^c$ by considering POVs on the adjacent side lanes with the lane-change decision generated through the module of Minimizing Overall Braking Induced by Lane Changes (MOBIL)~\cite{kesting2007general}. The detailed formulation of IDM and MOBIL, the hyper-parameters of the mentioned modules, and the simulation implementation in Python language can all be found at~\cite{highway_env}. 

In this study, we consider the testing performed within a three-lane straight-road (with 3.7 m lane-width) configuration involving the interaction between one subject vehicle and one POV persistently staying in front of the SV as shown in Fig.~\ref{fig:ads_oss_exp}(a). This is a common configuration in both related research work~\cite{feng2021intelligent,feng2020testing,weng2021towards,weng2022finite} and established vehicle safety testing standards~\cite{van2017euro,najm2007pre,forkenbrock2015nhtsa}. The OSS is thus a four-dimensional space as $S = D_x \times D_y \times V_0 \times V_1$ with the distance headway $d_x \in D_x = [0,50]$ m (i.e. the longitudinal offset between the center of the front bumper of the SV and the center of the rear bumper of the POV), the lateral offset $d_y \in D_y= [-3.7,3.7]$ m (i.e. the lateral distance between the geometric center of two vehicles), the set of global speeds of SV and the POV satisfying $V_0 = V_1 = [0, 25]$ m/s. Note that the distance headway $d_x$ is capped off at a sufficiently large value (50 m) as states associated with larger values are not of safety concern. The vehicle heading angle is not considered as the primary feature of OSS as it is often limited with the feasible maneuvers (lane-keeping and single lane change) in the given OSS. Similar to the Bipedal Locomotion case, some other states are considered as disturbances and uncertainties following a certain unknown distribution. Finally, the failure set $\mathcal{C}$ denotes the rear-end vehicle-to-vehicle collision event.

Note that the class of safety testing algorithms $\mathcal{TE}^{\pi}$ is considered for this case study, as the explicit testing policy is a relatively feasible and commonly observed methodology in the vehicle testing field. In particular, we consider five different testing policies controlling the motion of the lead POV denoted as $\pi_1^s, \pi_1^b, \pi_1^h, \pi_1^p$, and $\pi_1^e$ explained in detail as follows.
\paragraph{The steady-state policy $\pi_1^s$} The POV stays in the initialized lane and travels at the initialized speed indefinitely. 
\paragraph{The persistent-braking policy $\pi_1^b$} This policy shares a similar lateral controller with $\pi_1^s$, but the longitudinal control is replaced with a consistent brake-to-stop maneuver. This particular testing policy is commonly observed at various formal testing standards and research publications for safety evaluation of longitudinal Advanced Driver Assist System (ADAS) modules~\cite{forkenbrock2015nhtsa,weng2022formal,van2017euro}. 
\paragraph{The hybrid testing policy $\pi_1^h$} This is also a policy inspired by established testing standards~\cite{najm2007pre,national2019traffic}, where the POV takes a hybrid testing approach. If the POV is initialized in front of the SV within the same lane, we have $\pi_1^h=\pi_1^b$. Otherwise, the POV will first reach the nearest adjacent lane beside the SV, executes a heuristic longitudinal controller that seeks to force the value of $\tau=\frac{d_y}{v_0-v_1}$ (also know as the time-to-collision value in the literature~\cite{lee1976theory}) to satisfy $\tau\in [0,2]$~s and $d_x \geq 2$ m. If the desired condition is fulfilled, $\pi_1^h$ then proceeds with a lane-change to the SV's lane and an immediate brake-to-stop after the lane-change is accomplished. 
\paragraph{The model predictive testing policy $\pi_1^p$} The forth testing policy, $\pi_1^p$, follows the online adversarial testing approach proposed in~\cite{capito2020modeled} where the POV constantly predicts the future motion trajectory of the SV through model insights and assumptions, and tracks the predictive trajectory to force the $\ell_2$-norm distance between the two vehicles to be sufficiently small. In the example with the lead POV initialized in the left adjacent lane of the SV (see Fig.~\ref{fig:ads_oss_exp}) and let SV be equipped with $\pi_1^a$, $\pi_1^p$ may make consecutive single lane-change maneuvers to stay in front of the SV. This makes it different from $\pi_1^s$ and $\pi_1^b$ as the lead POV does not change lanes, or $\pi_1^h$ as the lead POV changes lanes at most once. One can refer to~\cite{capito2020modeled} for other examples of $\pi_1^h$ in safety testing of vehicles.
\paragraph{The learning-based testing policy with evolution strategy $\pi_1^e$} $\pi_1^e(\cdot;\theta)$ is configured as a two-layer neural network parameterized on $\theta$. The parameters are learnt through the evolution strategy with the reward design encourages rear-end collisions and short vehicle-to-vehicle distances. $\pi_1^e$ then uses the empirically best found parameters $\theta^*$ for all tests. The learning method is the same with the one adopted to learn the HZDES policy for Cassie in the last case study section. Given the state information of both vehicles, $\pi_1^e$ determines a desired speed and one of the three desired lateral control modes (lane-keeping, single lane-change to the left, and to the right) at 10 Hz. The desired speed is then tracked through the same IDM used by $\pi_0^c$. The selected lateral control mode is first mapped to a lateral target position and then tracked by another lateral PD controller. The PD controllers ensure the overall trajectory is smooth and feasible. 

Note that both $\pi_1^p$ and $\pi_1^e$ admit the same lane change gap-acceptance with $\pi_1^h$ as 2 m in respect to the common sense of responsibility. That is, the POV will reject to move to the SV's lane if the longitudinal distance headway $d_x \leq 2$. 
Moreover, some technical details of the testing policies are not discussed here as the primary focus of this paper is not to propose a particular adversarial testing policy, $\pi_1$, for the vehicle decision-making modules. The aforementioned five policies are selected as they are inspired by different sources (regulatory standards and research literature), they are created in different forms (concrete open-loop strategy, simple feedback policy, and complex neural networks), and as we should show later, they perform differently. One can refer to Fig.~\ref{fig:ads_oss_exp}(b) to (d) for some examples of behavioral differences among the mentioned testing policies. 
\begin{figure}
    \centering
    \includegraphics[width=0.49\textwidth]{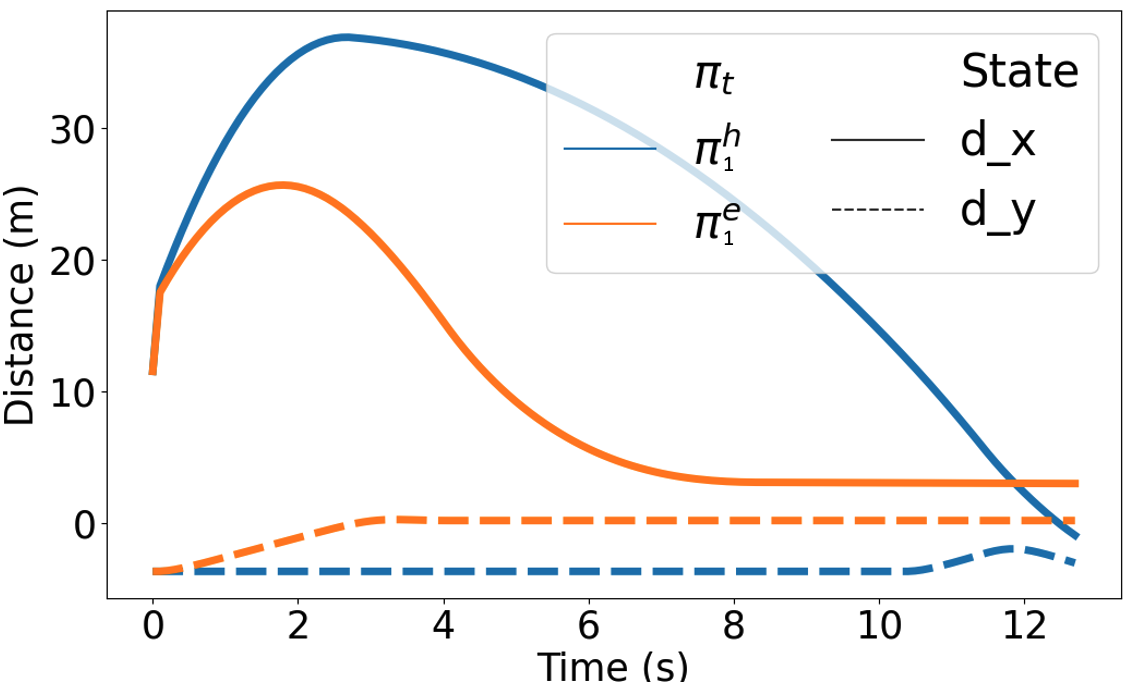}
    \caption{Given $\s_0$ as $d_x=10, d_y=-3.7, v_0=5$, and $v_1=20$, part of the state trajectories testing $\pi_0^a$ using $\pi_1^h$ and $\pi_1^e$ are compared. Note that $\pi_1^h$ manages to cause a collision and $\pi_1^e$ forces the SV to brake-to-stop without any failure event. }
    \label{fig:hve}
    \vspace{-5mm}
\end{figure}

\begin{table*}
    \centering
    \caption{The set difference analysis comparing all pairs of testing policies against the same decision-making module.}
    \label{tab:diversity}
    \resizebox{0.99\textwidth}{!}{%
    \begin{tabular}{l|l|c|c|c|c|c|c|c|c|c|c}
    \hline
    \multirow{2}{*}{$\pi_0$} &
    $\Phi_0$                        & $\Phi_s$  & $\Phi_s$  & $\Phi_s$  & $\Phi_s$  & $\Phi_b$  & $\Phi_b$  & $\Phi_b$  & $\Phi_h$    & $\Phi_h$  & $\Phi_p$ \\
    &$\Phi_1$                        & $\Phi_b$  & $\Phi_h$  & $\Phi_p$  & $\Phi_e$  & $\Phi_h$  & $\Phi_p$  & $\Phi_e$  & $\Phi_p$    & $\Phi_e$  & $\Phi_e$ \\\hline
    \multirow{2}{*}{$\pi_0^a$} & \texttt{IoU}($\Phi_0, \Phi_1$)  & 0.988   & 0.87 & 0.812 & 0.617 & 0.88 & 0.822 & 0.624 & 0.804   & 0.574 & 0.638 \\
    &$|\Phi_0 \setminus \Phi_1|/|O|$ & 0.012 & 0.125 & 0.181 & 0.368 & 0.114 & 0.169 & 0.356 & 0.116 & 0.315 & 0.245\\
    &$|\Phi_1 \setminus \Phi_0|/|O|$ & 0.0   & 0.0   & 0.0   & 0.0   & 0.0   & 0.0   & 0.0   & 0.06 & 0.07 & 0.058\\\hline
    \multirow{2}{*}{$\pi_0^c$} & \texttt{IoU}($\Phi_0, \Phi_1$)  & 0.959   & 0.866 & 0.832 & 0.623 & 0.903 & 0.868 & 0.645 & 0.871   & 0.601 & 0.637 \\
    &$|\Phi_0 \setminus \Phi_1|/|O|$ & 0.04 & 0.127 & 0.16 & 0.359 & 0.088 & 0.12 & 0.322 & 0.072 & 0.292 & 0.252\\
    &$|\Phi_1 \setminus \Phi_0|/|O|$ & 0.0   & 0.0   & 0.0   & 0.0   & 0.0   & 0.0   & 0.002   & 0.04 & 0.06 & 0.053\\\hline
    \end{tabular}%
    }
\end{table*}
\begin{table}[ht]
    \centering
    \caption{The empirically observed failure rate for $1000$ runs of scenarios with the initialization condition uniformly distributed in the given $O$.}
    \label{tab:fr}
    \resizebox{0.49\textwidth}{!}{%
    \begin{tabular}{l|c|c|c|c|c}
    \hline
    Testing Policy & $\pi_1^s$    & $\pi_1^b$   & $\pi_1^h$   & $\pi_1^p$   & $\pi_1^e$ \\ \hline
    Failure Rate with $\pi_0^a$ & 0.039    &0.051  &0.164  & 0.220 & 0.407 \\ \hline
    Failure Rate with $\pi_0^c$ & 0.05    &0.09  &0.178  & 0.213 & 0.41 \\ \hline
    \end{tabular}%
    }
\end{table}
For each pair of the decision-making module and testing policy $(\pi_0^x, \pi_1^y)$ with $x \in \{c, a\}$ and $y \in \{s, b, h, p, e\}$, one obtains an $\epsilon\delta$-almost safe set $\Phi_y^x$ with $\epsilon=0.01$, $\beta=0.0001$, and $\delta=[5, 3.7, 5, 5]$. Some examples of the sub-space slices for $\Phi_h^a$ and $\Phi_e^a$ are shown in Fig.~\ref{fig:phi_h} and Fig.~\ref{fig:phi_e}, respectively. Although $|\Phi_e^a|<|\Phi_h^a|$, note that each testing policy also manages to force the SV into some unsafe states that the other policy ends up with safe justifications, i.e. $|\Phi_e^a\setminus \Phi_h^a| \neq |\Phi_h^a \setminus \Phi_e^a| \neq 0$. For example, consider the lead POV starts from the left side lane near the SV at 10 meters ahead and operates at 15 m/s faster than the SV with the decision-making module of $\pi_0^a$ (i.e. near the upper right corner of the first column, second row sub-plot in Fig.~\ref{fig:phi_h} and Fig.~\ref{fig:phi_e}), the obtained runs of scenarios are shown in Fig.~\ref{fig:hve} against $\pi_1^h$ and $\pi_1^e$. Note that $\pi_1^h$ stayed on the side lane significantly longer than $\pi_1^e$ (induced by the non-zero lateral offset) and executed the lane-change and braking maneuver at the appropriate moment to force the collision. On the other hand, $\pi_1^e$ started the attack (lane change and braking) too early and too abrupt, failed to force the failure outcome before the SV braking-to-stop. The above observation is very common among the studied pairs of testing algorithms as illustrated in Table~\ref{tab:diversity}. Note that the intersection-over-union (IoU) ratio \texttt{IoU}($\Phi_0, \Phi_1$) is defined as $\frac{|\Phi_0\cap \Phi_1|}{|\Phi_0 \cup \Phi_1|}$. As the IoU ratio of all pairs of almost safe sets are smaller than one, all sets are different. Furthermore, note that the non-zero value of $|\Phi_0 \setminus \Phi_1|/|O|$ implies that $\Phi_0$ contains some states that are not covered by $\Phi_1$. Among the testing policy pairs involving $\pi_1^h$, $\pi_1^p$, and $\pi_1^e$, the set difference exists in both ways for both subject policies (as shown by the last three columns of Table~\ref{tab:diversity}). That is, there are always some states initialized from which one testing policy forces unsafe outcomes while the other policy fails to do so. On the other hand, note that for both $x\in\{a,c\}$, $\Phi_s^x \subset \Phi_b^x \subset \Phi_p^x$, $\Phi_b^x \subset \Phi_h^x$ and $\Phi_b^a \subset \Phi_e^a$. By Definition~\ref{def:agg}, the persistent-braking policy $\pi_1^b$ is $\epsilon$-almost more adversarial than the steady-state policy $\pi_1^s$, and the hybrid testing policy $\pi_1^h$ is $\epsilon$-almost more adversarial than the persistent-braking policy $\pi_1^b$, at the given resolution level dictated by $\delta$. 

A further exploration of Algorithm~\ref{alg:agg_val} ends up with the same outcome identified by the above quantification outcomes. Note that for policy pairs that are significantly different (e.g. $(\pi_1^e, \pi_1^s)$ and $(\pi_1^e, \pi_1^s)$), even if the SV survives the desired runs of scenarios, one rarely satisfies~\eqref{eq:agg_val3} and line 14 of Algorithm~\ref{alg:agg_val} is thus triggered with the almost safe set quantification of both testing policies. If tested against $\pi_0$ uniformly distributed in $\{\pi_0^a, \pi_0^c\}$ with $\epsilon=0.01, \beta=0.0001, \delta=[10,3.7,5,5]$, among the 20 pairwise permutations of the five testing policies ($\frac{5!}{(5-2)!}$), half of them end up satisfying line 12 in Algorithm~\ref{alg:agg_val} with exactly 917 runs of scenarios for each validation (by~\eqref{eq:agg_val1}), 4 of them trigger line 14 for the complete quantification with 2398 runs of scenarios on average. The rest of the pairs terminate at line 6 with an observed collision. Similar to the Cassie case with Table~\ref{tab:efficiency}, the sampling efficiency can be further improved for larger $\epsilon$, $\delta$-neighbourhood, and $\beta$.

Note that the above subtle differences among testing approaches are made possible by the set-based safety metric presented in this paper. If adopting the traditional observed risk-base approach, as shown in Table~\ref{tab:fr}, the aggressiveness of the testing policy $\pi_1^y$ grows monotonically w.r.t. the order of $\{s, b, h, p, e\}$. The learned policy from evolution strategy is thus the most aggressive approach for both subject policies. However, recall the previous discussion, it is not necessarily the case as the basic transitive property is not satisfied (e.g. the SV tested safe against $\pi_1^e$ in the case shown in Fig.~\ref{fig:hve} is not safe against $\pi_1^h$).

We conclude this section by emphasizing that the absolute black-box assumption by Remark~\ref{rmk:bbox} is technically impossible for the specific OSS studied in this section even if one assumes the subject vehicle's decision-making module being a black-box. Due to the powertrain dynamics, one cannot request an arbitrary acceleration at any time. Thanks to the non-holonomic nature of vehicles, the lateral motion is strictly limited. Given the commonly agreed traffic rules, a vehicle typically stays persistently stationary after brake-to-stop. These constraints and possibly other limitations are established through over a century (since the first invention of the vehicle in 1886) of joint efforts across various entities, regions, and research disciplines. As a result, Theorem~\ref{thm:nfl} is not valid as the testing system is not a black-box in its complete OSS. However, the black-box nature may still hold in a certain subset of OSS, and the implications of the impossibility theorem of black-box testing is still there. As we have demonstrated above without a particular testing policy dominating the safety test outcomes at all states. Moreover, it remains possible that Theorem~\ref{thm:nfl} applies to a certain other OSS where Remark~\ref{rmk:bbox} is valid, yet details are beyond the scope of this paper.

\section{Conclusion}\label{sec:conclusion}
This paper studies the class of black-box scenario-based safety testing algorithm with a particular focus on the formal justification of the algorithm's aggressiveness. For algorithms of the Type-I difference with a variety of different state-action exploration orders, an impossibility theorem for safety testing (Theorem~\ref{thm:nfl}) is presented, which provably reveals that all algorithms are equal in performance, hence the aggressiveness comparability does not exist in the studied case. Moreover, for algorithms of the Type-II difference with different sets of state-dependent testing actions and feedback testing policies being adopted, the presented impossibility theorem no longer applies. We thus propose the idea of using the $\epsilon\delta$-almost (controlled) safe set to help characterizing the performance of the testing method (with Definition~\ref{def:agg_pu} and Definition~\ref{def:agg}). A practically efficient (as discussed empirically in Section~\ref{sec:case}) and provably unbiased (by Theorem~\ref{thm:agg}) algorithm is also presented that helps justifying if a given testing approach is more adversarial/aggressive than the other. Finally, the case study with bipedal locomotion controllers and vehicle decision-making modules are also presented. The empirical observations support various theoretical claims presented in this paper.

For algorithms of Type-I difference, it is of future interest to expand the implications of the presented impossibility theorem with more challenging cases. For algorithms of Type-II difference, especially with the Algorithm~\ref{alg:agg_val}, the curse-of-dimensionality requires some future attention as the algorithm can be computational inefficient with high-dimensional OSS designs. The algorithmic differences can be further extended to a combination of Type-I and Type-II, as well as other possible forms. The aggressiveness analysis remains a challenge for those complex extensions.

\appendices

\section{}\label{apx:notation}

\nomenclature{$\s$}{Testing system state}
\nomenclature{$S$}{State space}
\nomenclature{$U$}{Testing action space}
\nomenclature{$U(\s)$}{State-dependent testing action space}
\nomenclature{$\pi$}{Testing policy}
\nomenclature{$\mathcal{C}$}{Set of failure states}
\nomenclature{$\uu$}{Testing action}
\nomenclature{$O$}{Operational state space (OSS)}
\nomenclature{$f_s$}{The discrete-time testing system dynamics}
\nomenclature{$F_s$}{The function space of all possible $f_s$}
\nomenclature{$\sigma(k)$}{ A scenario of $k$ steps}
\nomenclature{$\mathcal{R}_{\sigma}(\s_0, k)$}{A run of a scenario initialized at $\s_0$}
\nomenclature{$\mathcal{R}(\s_0)$}{$\mathcal{R}_{\sigma}(\s_0, k)$ assuming all scenarios defined over the same time domain}
\nomenclature{$\mathcal{TE}$}{The class of scenario-based safety evaluation algorithms}
\nomenclature{$\mathcal{TE}^{\pi}$}{$\mathcal{TE}$ with actions generated from a given $\pi$}
\nomenclature{$\mathcal{TE}^{u}$}{$\mathcal{TE}$ with open-loop state-action exploration through sampling or brute-force}
\nomenclature{$c$}{Safety cost function for a run of a scenario}
\nomenclature{$\mathcal{T}$}{The termination function for $\mathcal{TE}$}
\nomenclature{$G$}{The target set of the cost function}
\nomenclature{$\bar{\uu}$}{A sequence of ($k$) actions for a certain $\mathcal{R}(\s_0)$}
\nomenclature{$g_m$}{A sequence of safety costs obtained from $m$ runs of scenarios}
\nomenclature{$\mathcal{N}_{\delta}(\s)$}{The (extended) $\delta$-neighborhood of $\s$}
\nomenclature{$\Phi_{\delta}^O$}{The extended $\delta$-covering set of $O$}
\nomenclature{$\Phi_{s}^O$}{The set of centroids for $\Phi_{\delta}^O$}
\nomenclature{$\Phi_{\delta}$}{The $\delta$-covering set of an implicit $O$}
\nomenclature{$\Phi_{s}$}{The set of centroids for $\Phi_{\delta}$}
\nomenclature{$\bar{F}_s$}{A subset of $F_s$}
\nomenclature{$\delta$}{The resolution coefficient for the $\delta$-covering set}
\nomenclature{$\epsilon$}{The probability coefficient for the almost safe set}
\nomenclature{$\beta$}{The desired confidence level coefficient for the almost safe set}
\nomenclature{$\pi_0$}{The subject-vehicle control policy for the vehicle-related testing applications}
\nomenclature{$\pi_1$}{The lead-vehicle control policy (i.e. the testing policy) for the vehicle-related testing applications}
\nomenclature{$w$}{Disturbances and uncertainties}
\nomenclature{$W$}{Set of disturbances and uncertainties}
\nomenclature{$k$}{Number of states in a run of a scenario}
\nomenclature{$m$}{Number of runs of scenarios}
\printnomenclature

\section{Proof of Theorem~\ref{thm:nfl}}\label{apx:nfl}
The following proof first transfers the notion of $\mathbb{P}(g_m \mid f_s, m, \mathcal{TE}^{u})$ given an arbitrary safety cost design to the same format studied by the previously proposed NFL theorems in the black-box optimization literature~\cite{wolpert1997no}. One then follows the similar technique from Appendix A in~\cite{wolpert1997no} to finish the proof. Note that the inner product formulas discussed in~\cite{wolpert2021important} potentially give rise to an alternative to the presented proof, yet details are beyond the scope of this paper. 
\begin{proof}
The propagation of a run of a scenario starting from $\s_0 \in O$ with $f_s$ against a sequence of selected testing actions $\bar{\uu}$ is simply a mean to acquire the required input of the cost function $c$, i.e. a run of a scenario. One can rewrite the required input scenario run as a function of the initialization state and the testing action sequence as 
\begin{equation}
    \begin{aligned}
        & \mathcal{R}(\s_0) \setminus \{\s_0\} = \begin{bmatrix} f_s(\s_0,\uu[1]) & f_s(f_s(\s_0,\uu[1]), \uu[2]) & \ldots \end{bmatrix}
        \\ & = \begin{bmatrix} \bar{f}_1(\s_0,\uu[1]) & \bar{f}_2(\s_0,[\uu[1],\uu[2]]) & \ldots & \bar{f}_{m-1}(\s_0,\uu) \end{bmatrix}
        \\ & = \bar{f}(\s_0, \bar{\uu})
    \end{aligned}
\end{equation}
and $\bar{f}: O \times U^{k} \rightarrow O^k$. Note that $\bar{f}$ is not uniformly distributed on $O^{k^{(O \times U^{k})}}$. However, as $f_s$ is uniformly distributed on $O^{O\times U}$ by Remark~\ref{rmk:bbox}, each function entry of $\bar{f}$, $\bar{f}_i, i \in \Z_{m-1}$, is uniformly distributed on $O_i^{V_i}$ for a certain $O_i \subset O, V_i \subset O\times U^{i}$ and $O_1 = O, V_1=O\times U$. As a result, we have $\bar{f}$ being uniformly distributed on $\bar{F} = \bar{O}^{\bar{V}}$ with $\bar{O} \subset O^k$ and $\bar{V} \subset O \times U^{k}$. Let $o_m \in \bar{O}^m$ be a sequence of $m$ runs of scenarios. 

The proof then follows with two steps. The first step is to show that the acquisition of any $o_m$ is irrelevant from the selected testing algorithm $\mathcal{TE}^u$, i.e.
\begin{equation}\label{eq:scene_nfl}
    \sum_{\bar{f} \in \bar{F}}\mathbb{P}(o_m \mid \bar{f}, m, \mathcal{TE}^{u}_1) = \sum_{\bar{f} \in \bar{F}}\mathbb{P}(o_m \mid \bar{f}, m, \mathcal{TE}^{u}_2)
\end{equation}
The second step is to associate $\mathbb{P}(o_m \mid \bar{f}, m, \mathcal{TE}^{u})$ with $\mathbb{P}(g_m \mid f_s, m, \mathcal{TE}^{u})$ in a way that is irrelevant from the testing algorithm $\mathcal{TE}^u$.

We start from the first step. Note that \eqref{eq:scene_nfl} is similar to a standard optimization form analyzed in the previous literature~\cite{wolpert1997no}. The overall proof then follows a two-step induction. One first considers $m = 1$, we have
\begin{equation}\label{eq:nfl_pf_1}
    \begin{aligned}
    & \sum_{\bar{f} \in \bar{F}}\mathbb{P}(o_1 \mid \bar{f}, m=1, \mathcal{TE}^{u}) \\
    & = \sum_{\bar{f} \in \bar{F}} \llbracket o_1, \bar{f}(\s_0^1, \bar{\uu}^1) \rrbracket = \frac{|\bar{V}|^{|\bar{O}||\bar{V}|}}{|\bar{V}|}.
    \end{aligned}
\end{equation}
The last equality is established as $\llbracket o_1, \bar{f}(\s_0^1, \bar{\uu}^1) \rrbracket=1$ only for the particular $\bar{f}$ that propagates the exact run of a scenario $o_1$ given $\s_0^1$ and $\bar{\uu}^1$.
The induction then follows with
\begin{equation}\label{eq:nfl_pf_m}
    \begin{aligned}
        & \sum_{\bar{f} \in \bar{F}} \mathbb{P}(o_{m+1} \mid \bar{f}, m+1, \mathcal{TE}^u) \\
        &= \sum_{\bar{f} \in \bar{F}}\mathbb{P}(\{o_{m+1}[1], \ldots, o_{m+1}[m+1]\} \mid \bar{f}, m+1, \mathcal{TE}^u) \\
        & = \sum_{\bar{f} \in \bar{F}} \mathbb{P}(o_{m+1}[m+1] \mid o_m, \bar{f}, m+1, \mathcal{TE}^u) \\& \ \ \ \  \cdot\mathbb{P}(o_m \mid \bar{f}, m+1, \mathcal{TE}^u) \\
        & = \frac{1}{|\bar{O}|}\sum_{\bar{f} \in \bar{F}}\mathbb{P}(o_{m} \mid \bar{f}, m+1, \mathcal{TE}^u) \\
        & = \frac{1}{|\bar{O}|}\sum_{\bar{f} \in \bar{F}}\mathbb{P}(o_{m} \mid \bar{f}, m, \mathcal{TE}^u).
    \end{aligned}
\end{equation}
Combining~\eqref{eq:nfl_pf_1} and \eqref{eq:nfl_pf_m} implies that $\sum_{\bar{f} \in \bar{F}}\mathbb{P}(o_m \mid \bar{f}, m, \mathcal{TE}^{u})$ is independent of $\mathcal{TE}^{u}$. 

To proceed with the second step, consider an arbitrary cost function $c: \bar{O} \rightarrow G$. For each $g_m\in G^m$ and $m \in \Z$, one must has a corresponding set of $J$ sequences of runs of scenarios $\{o_m^j\}_{j \in \Z_J}$ such that the obtained sequence of costs remain as $g_m$ for all $o_m \in \{o_m^j\}_{j \in \Z_J}$. Therefore,
\begin{equation}
    \sum_{f_s \in F_s}\mathbb{P}(g_m \mid f_s, m, \mathcal{TE}^{u}) = \sum_{j\in \Z_J} \sum_{\bar{f} \in \bar{F}}\mathbb{P}(o_m^j \mid \bar{f}, m, \mathcal{TE}^{u}).
\end{equation}
As the right hand side of the above equation is irrelevant of the testing algorithm $\mathcal{TE}_u$, so is the left hand side. This completes the proof.
\end{proof}

\section{Proof of Theorem~\ref{thm:agg}}\label{apx:agg}
\begin{proof}
If $\Phi'_1\setminus \Phi_1 = \emptyset$, the two policies are immediately equally aggressive by Theorem~\ref{thm:validation} and Definition~\ref{def:agg}.

Otherwise, let
\begin{equation}\label{eq:p_star}
    p* = \mathbb{P}(\{\s \in \Phi'_1\setminus \Phi_1, \mathcal{R}(\s) \cap \mathcal{C}\neq \emptyset \}),
\end{equation}
we consider the following two cases. 

(i) If $p^* < \epsilon$, by Definition~\ref{def:epsilon-almost-ss} the two policies are equally aggressive. 

(ii) Otherwise ($p^* \geq \epsilon$), we have
\begin{equation}
    \sum_{\s \in \Phi'_1 \setminus \Phi_1} p_S(\s) \cdot \mathbb{P}(\{\mathcal{R}(\s) \cap \mathcal{C}\neq \emptyset \}) < \epsilon
\end{equation}
by~\eqref{eq:agg_val3} and \eqref{eq:p_star}. 

That is, there must exist $\Phi_2 \supseteq \Phi'_1$ and $\Phi_2$ is $\epsilon$-almost safe with confidence level at least $1-\beta$. This proves the theorem.
\end{proof}

\section{The Almost Safe Set Quantification Algorithm}\label{apx:qnt}

The almost safe set quantification algorithm $\mathcal{QNT}_{\mathcal{TE}^{\pi}}(O, \bar{F}_s, \epsilon, \beta, \delta)$ is summarized in Algorithm~\ref{alg:qnt} adapted from~\cite{weng2022leeged} using the notations presented in this paper. Note that \texttt{pop}, \texttt{reachable}, \texttt{norm-nearest}, \texttt{remove}, and \texttt{append} are all notional functions. $\mathcal{X}.$\texttt{pop}() returns a point $\x\in\mathcal{X}$ and removes it from the set. \texttt{reachable} ($\s, G$) returns all vertices on the graph $G$ that connects, directly and indirectly, to the point $\s$ through a depth-first-search routine. The commands \texttt{remove} and \texttt{append} simply remove a point from or add a point to the given set, respectively. $\mathcal{X}$.\texttt{norm-nearest}($\x$) returns the nearest point to $\x$ in $\mathcal{X}$ in terms of the normalized $\ell_2$-norm distance. That is, $\x$ and all points in $\mathcal{X}$ are first normalized w.r.t. the admissible value range of each individual dimension and one then propagates the $\ell_2$-norm distance between the normalized $\x$ and $\mathcal{X}$. 

Algorithm~\ref{alg:qnt} can be directly applied to the case study with decision-making modules. To apply the algorithm to the bipedal locomotion controller case, one should (i) replace line 9 with uniform sampling of the external force from $F_x \times F_y$ for the state-invariant $U$ (e.g. Fig.~\ref{fig:cassie_uts}) or (ii) make the state-dependent $U(\s)$ part of the given OSS (i.e. replace the input OSS $O$ with $O' = \cup_{\s \in O} \{\s\} \times U(\s)$) and the run of a scenario at line 9 simply follows the selected initialization state from the modified set $O'$. 

\begin{algorithm}[H]
\small
    \begin{algorithmic}[1]
    \State {\bf Input:} $O$, $\bar{F}_s$, $\mathcal{C}$, $\epsilon$, $\beta$, $\delta$, $\pi$ from $\mathcal{TE}^{\pi}$.
    \State {\bf Initialize: } $\delta$-covering set $\Phi_{\delta}=\Phi_{\delta}^O$ with centroids $\Phi_s$, graph $G_{s} = (\Phi_s, E_s), E_s =\emptyset
    \subset O^2$ and $G_u=(D_u, E_u), D_u=\emptyset\subset S, E_u = \emptyset \subset S^2$, prioritized replay buffer $\mathcal{B}=\emptyset$, N=0.
    \State{{\bf While} $N<\frac{\ln{\beta}}{\ln{(1-\epsilon)}}$:}
    \State{\ \ \ \ {\bf If} $\mathcal{B}=\emptyset$}
    \State{\ \ \ \ \ \ \ \ $\s_0 \sim \textit{U}(\Phi_s)$}
    \State{\ \ \ \ {\bf Else}}
    \State{\ \ \ \ \ \ \ \ $\s_b = \mathcal{B}$.\texttt{pop}(), $\s_0 = \Phi_s.$\texttt{norm-nearest}($\s_b$)}
    \State{\ \ \ \ {\bf End If}}
    \State{\ \ \ \ Get $\tau=\mathcal{R}(\s_0)$ with $f_s \sim \textit{U}(\bar{F}_s)$ and $\pi$}
    \State{\ \ \ \ {\bf If} $\tau\cap\mathcal{C} \neq \emptyset$}
    \State{\ \ \ \ \ \ \ \ {\bf For $i$ in $\Z_{|\tau|-1}$} {\bf do}}
    \State{\ \ \ \ \ \ \ \ \ \ \ \ $\mathcal{B}$.\texttt{append}($\tau[i]$)}
    \State{\ \ \ \ \ \ \ \ \ \ \ \ {\bf For} $\s$ in \texttt{Reachable}($\tau[i],G_s)$ {\bf do}}
    \State{\ \ \ \ \ \ \ \ \ \ \ \ \ \ \ \ $\Phi_s$.\texttt{remove}($\s$)}
    \State{\ \ \ \ \ \ \ \ \ \ \ \ $E_u$.\texttt{append}(($\tau[i]$, $\tau[i+1]$))}
    \State{\ \ \ \ \ \ \ \ \ \ \ \ {\bf End For}}
    \State{\ \ \ \ \ \ \ \ $\mathcal{B}$.\texttt{append}($\tau[i+1]$)}
    \State{\ \ \ \ \ \ \ \ {\bf End For}}
    \State{\ \ \ \ \ \ \ \ $N=0$}
    \State{\ \ \ \ {\bf Else}}
    \State{\ \ \ \ \ \ \ \ $\bar{\s}=\s_0, N_s=|\Phi_s|$}
    \State{\ \ \ \ \ \ \ \ {\bf For $i$ in $\{2,\ldots,|\tau|\}$} {\bf do}}
    \State{\ \ \ \ \ \ \ \ \ \ \ \ {\bf If} $\tau[i] \notin \Phi_{\delta}$}
    \State{\ \ \ \ \ \ \ \ \ \ \ \ \ \ \ \ $E_s$.\texttt{append}(($\bar{\s}$, $\tau[i]$))}
    \State{\ \ \ \ \ \ \ \ \ \ \ \ \ \ \ \ $\bar{\s}=\tau[i]$}
    \State{\ \ \ \ \ \ \ \ \ \ \ \ {\bf End If}}
    \State{\ \ \ \ \ \ \ \ {\bf End For}}
    \State{\ \ \ \ \ \ \ \ {\bf If} $N_s=|\Phi_s|$ and $\mathcal{B}=\emptyset$}
    \State{\ \ \ \ \ \ \ \ \ \ \ \ $N += 1$}
    \State{\ \ \ \ \ \ \ \ {\bf Else}}
    \State{\ \ \ \ \ \ \ \ \ \ \ \ $N = 0$}
    \State{\ \ \ \ \ \ \ \ {\bf End If}}
    \State{\ \ \ \ {\bf End If}}
    \State {{\bf Output:} $\Phi_{\delta}, \Phi_s$}
    \end{algorithmic}
    \caption{Almost Safe Set Quantification} \label{alg:qnt}
\end{algorithm}
The example code of Algorithm~\ref{alg:qnt} in Python can be found at~\cite{sdq_tools}.

\bibliographystyle{IEEEtran}
\bibliography{output}

\begin{thebibliography}{10}
\providecommand{\url}[1]{#1}
\csname url@samestyle\endcsname
\providecommand{\newblock}{\relax}
\providecommand{\bibinfo}[2]{#2}
\providecommand{\BIBentrySTDinterwordspacing}{\spaceskip=0pt\relax}
\providecommand{\BIBentryALTinterwordstretchfactor}{4}
\providecommand{\BIBentryALTinterwordspacing}{\spaceskip=\fontdimen2\font plus
\BIBentryALTinterwordstretchfactor\fontdimen3\font minus
  \fontdimen4\font\relax}
\providecommand{\BIBforeignlanguage}[2]{{%
\expandafter\ifx\csname l@#1\endcsname\relax
\typeout{** WARNING: IEEEtran.bst: No hyphenation pattern has been}%
\typeout{** loaded for the language `#1'. Using the pattern for}%
\typeout{** the default language instead.}%
\else
\language=\csname l@#1\endcsname
\fi
#2}}
\providecommand{\BIBdecl}{\relax}
\BIBdecl

\bibitem{riedmaier2020survey}
S.~Riedmaier, T.~Ponn, D.~Ludwig, B.~Schick, and F.~Diermeyer, ``Survey on
  scenario-based safety assessment of automated vehicles,'' \emph{IEEE Access},
  vol.~8, pp. 87\,456--87\,477, 2020.

\bibitem{lee1976theory}
D.~N. Lee, ``A theory of visual control of braking based on information about
  time-to-collision,'' \emph{Perception}, vol.~5, no.~4, pp. 437--459, 1976.

\bibitem{wishart2020driving}
J.~Wishart, S.~Como, M.~Elli, B.~Russo, J.~Weast, N.~Altekar, E.~James, and
  Y.~Chen, ``Driving safety performance assessment metrics for ads-equipped
  vehicles,'' \emph{SAE Technical Paper}, vol.~2, no. 2020-01-1206, 2020.

\bibitem{manuele2009leading}
F.~A. Manuele, ``Leading \& lagging indicators,'' \emph{Professional Safety},
  vol.~54, no.~12, p.~28, 2009.

\bibitem{weng2022finite}
B.~Weng, L.~Capito, U.~Ozguner, and K.~Redmill, ``A finite-sampling,
  operational domain specific, and provably unbiased connected and automated
  vehicle safety metric,'' \emph{IEEE Transactions on Intelligent
  Transportation Systems}, pp. 1--13, 2022.

\bibitem{zhao2016accelerated}
D.~Zhao, H.~Lam, H.~Peng, S.~Bao, D.~J. LeBlanc, K.~Nobukawa, and C.~S. Pan,
  ``Accelerated evaluation of automated vehicles safety in lane-change
  scenarios based on importance sampling techniques,'' in \emph{IEEE
  Transactions on Intelligent Transportation Systems}, vol.~18, no.~3.\hskip
  1em plus 0.5em minus 0.4em\relax IEEE, 2016, pp. 595--607.

\bibitem{feng2021intelligent}
S.~Feng, X.~Yan, H.~Sun, Y.~Feng, and H.~X. Liu, ``Intelligent driving
  intelligence test for autonomous vehicles with naturalistic and adversarial
  environment,'' \emph{Nature Communications}, vol.~12, no.~1, pp. 1--14, 2021.

\bibitem{ding2022survey}
W.~Ding, C.~Xu, H.~Lin, B.~Li, and D.~Zhao, ``A survey on safety-critical
  scenario generation from methodological perspective,'' \emph{arXiv preprint
  arXiv:2202.02215}, 2022.

\bibitem{feng2023dense}
S.~Feng, H.~Sun, X.~Yan, H.~Zhu, Z.~Zou, S.~Shen, and H.~X. Liu, ``Dense
  reinforcement learning for safety validation of autonomous vehicles,''
  \emph{Nature}, vol. 615, no. 7953, pp. 620--627, 2023.

\bibitem{koren2018adaptive}
M.~Koren, S.~Alsaif, R.~Lee, and M.~J. Kochenderfer, ``Adaptive stress testing
  for autonomous vehicles,'' in \emph{2018 IEEE Intelligent Vehicles Symposium
  (IV)}.\hskip 1em plus 0.5em minus 0.4em\relax IEEE, 2018, pp. 1--7.

\bibitem{tuncali2018simulation}
C.~E. Tuncali, G.~Fainekos, H.~Ito, and J.~Kapinski, ``Simulation-based
  adversarial test generation for autonomous vehicles with machine learning
  components,'' in \emph{2018 IEEE Intelligent Vehicles Symposium (IV)}.\hskip
  1em plus 0.5em minus 0.4em\relax IEEE, 2018, pp. 1555--1562.

\bibitem{klischat2019generating}
M.~Klischat and M.~Althoff, ``Generating critical test scenarios for automated
  vehicles with evolutionary algorithms,'' in \emph{2019 IEEE Intelligent
  Vehicles Symposium (IV)}.\hskip 1em plus 0.5em minus 0.4em\relax IEEE, 2019,
  pp. 2352--2358.

\bibitem{kuutti2020training}
S.~Kuutti, S.~Fallah, and R.~Bowden, ``Training adversarial agents to exploit
  weaknesses in deep control policies,'' in \emph{2020 IEEE International
  Conference on Robotics and Automation (ICRA)}.\hskip 1em plus 0.5em minus
  0.4em\relax IEEE, 2020, pp. 108--114.

\bibitem{feng2020testing}
S.~Feng, Y.~Feng, C.~Yu, Y.~Zhang, and H.~X. Liu, ``Testing scenario library
  generation for connected and automated vehicles, part i: Methodology,''
  \emph{IEEE Transactions on Intelligent Transportation Systems}, 2020.

\bibitem{li2020av}
G.~Li, Y.~Li, S.~Jha, T.~Tsai, M.~Sullivan, S.~K.~S. Hari, Z.~Kalbarczyk, and
  R.~Iyer, ``{AV-FUZZER}: Finding safety violations in autonomous driving
  systems,'' in \emph{2020 IEEE 31st International Symposium on Software
  Reliability Engineering (ISSRE)}.\hskip 1em plus 0.5em minus 0.4em\relax
  IEEE, 2020, pp. 25--36.

\bibitem{ding2021multimodal}
W.~Ding, B.~Chen, B.~Li, K.~J. Eun, and D.~Zhao, ``Multimodal safety-critical
  scenarios generation for decision-making algorithms evaluation,'' \emph{IEEE
  Robotics and Automation Letters}, vol.~6, no.~2, pp. 1551--1558, 2021.

\bibitem{innes2021automated}
C.~Innes and S.~Ramamoorthy, ``Automated testing with temporal logic
  specifications for robotic controllers using adaptive experiment design,'' in
  \emph{2022 International Conference on Robotics and Automation (ICRA)}.\hskip
  1em plus 0.5em minus 0.4em\relax IEEE, 2022, pp. 6814--6821.

\bibitem{hussain2018lteinspector}
S.~Hussain, O.~Chowdhury, S.~Mehnaz, and E.~Bertino, ``Lteinspector: A
  systematic approach for adversarial testing of 4g lte,'' in \emph{Network and
  Distributed Systems Security (NDSS) Symposium 2018}, 2018.

\bibitem{capito2020modeled}
L.~Capito, B.~Weng, U.~Ozguner, and K.~Redmill, ``A modeled approach for online
  adversarial test of operational vehicle safety,'' in \emph{2021 American
  Control Conference (ACC)}.\hskip 1em plus 0.5em minus 0.4em\relax IEEE, 2021,
  pp. 398--404.

\bibitem{najm2007pre}
W.~G. Najm, J.~D. Smith, M.~Yanagisawa \emph{et~al.}, ``Pre-crash scenario
  typology for crash avoidance research,'' United States. National Highway
  Traffic Safety Administration, Tech. Rep., 2007.

\bibitem{forkenbrock2015nhtsa}
G.~J. Forkenbrock and A.~S. Snyder, ``{NHTSA}’s 2014 automatic emergency
  braking test track evaluations,'' National Highway Traffic Safety
  Administration, Tech. Rep., 2015.

\bibitem{van2017euro}
M.~R. van Ratingen, ``The euro ncap safety rating,'' in \emph{Karosseriebautage
  Hamburg 2017}, A.~Piskun, Ed.\hskip 1em plus 0.5em minus 0.4em\relax
  Wiesbaden: Springer Fachmedien Wiesbaden, 2017, pp. 11--20.

\bibitem{national2019traffic}
N.~H. T.~S. Administration \emph{et~al.}, ``Traffic jam assist system
  confirmation test (working draft),'' 2019.

\bibitem{winner2019pegasus}
H.~Winner, K.~Lemmer, T.~Form, and J.~Mazzega, ``Pegasus—first steps for the
  safe introduction of automated driving,'' in \emph{Road Vehicle Automation
  5}.\hskip 1em plus 0.5em minus 0.4em\relax Springer, 2019, pp. 185--195.

\bibitem{madry2018towards}
A.~Madry, A.~Makelov, L.~Schmidt, D.~Tsipras, and A.~Vladu, ``Towards deep
  learning models resistant to adversarial attacks,'' \emph{stat}, vol. 1050,
  p.~9, 2017.

\bibitem{corso2019adaptive}
A.~Corso, P.~Du, K.~Driggs-Campbell, and M.~J. Kochenderfer, ``Adaptive stress
  testing with reward augmentation for autonomous vehicle validation,'' in
  \emph{2019 IEEE Intelligent Transportation Systems Conference (ITSC)}.\hskip
  1em plus 0.5em minus 0.4em\relax IEEE, 2019, pp. 163--168.

\bibitem{wang2021interaction}
X.~Wang, H.~Peng, S.~Zhang, and K.-H. Lee, ``An interaction-aware evaluation
  method for highly automated vehicles,'' in \emph{2021 IEEE International
  Intelligent Transportation Systems Conference (ITSC)}.\hskip 1em plus 0.5em
  minus 0.4em\relax IEEE, 2021, pp. 394--401.

\bibitem{barcelo2003safety}
J.~Barcel{\'o}~Bugeda, A.-G. Dumont, L.~Montero~Mercad{\'e}, J.~Perarnau, and
  A.~Torday, ``Safety indicators for microsimulation-based assessments,'' in
  \emph{Transportation Research Board 82nd Annual Meeting}.\hskip 1em plus
  0.5em minus 0.4em\relax TRB, 2003, pp. 1--18.

\bibitem{bellem2018comfort}
H.~Bellem, B.~Thiel, M.~Schrauf, and J.~F. Krems, ``Comfort in automated
  driving: An analysis of preferences for different automated driving styles
  and their dependence on personality traits,'' \emph{Transportation research
  part F: traffic psychology and behaviour}, vol.~55, pp. 90--100, 2018.

\bibitem{wolpert1997no}
D.~H. Wolpert and W.~G. Macready, ``No free lunch theorems for optimization,''
  \emph{IEEE transactions on evolutionary computation}, vol.~1, no.~1, pp.
  67--82, 1997.

\bibitem{wolpert2021important}
D.~H. Wolpert, ``What is important about the no free lunch theorems?'' in
  \emph{Black Box Optimization, Machine Learning, and No-Free Lunch
  Theorems}.\hskip 1em plus 0.5em minus 0.4em\relax Springer, 2021, pp.
  373--388.

\bibitem{lechner2021adversarial}
M.~Lechner, R.~Hasani, R.~Grosu, D.~Rus, and T.~A. Henzinger, ``Adversarial
  training is not ready for robot learning,'' in \emph{2021 IEEE International
  Conference on Robotics and Automation (ICRA)}.\hskip 1em plus 0.5em minus
  0.4em\relax IEEE, 2021, pp. 4140--4147.

\bibitem{hauer2020re}
F.~Hauer, A.~Pretschner, and B.~Holzm{\"u}ller, ``Re-using concrete test
  scenarios generally is a bad idea,'' in \emph{2020 IEEE Intelligent Vehicles
  Symposium (IV)}.\hskip 1em plus 0.5em minus 0.4em\relax IEEE, 2020, pp.
  1305--1310.

\bibitem{weng2021towards}
B.~Weng, L.~Capito, U.~Ozguner, and K.~Redmill, ``Towards guaranteed safety
  assurance of automated driving systems with scenario sampling: An invariant
  set perspective,'' \emph{IEEE Transactions on Intelligent Vehicles}, vol.~7,
  no.~3, pp. 638--651, 2021.

\bibitem{weng2022formal}
B.~Weng, M.~Zhu, and K.~Redmill, ``A formal safety characterization of advanced
  driver assist systems in the car-following regime with scenario-sampling,''
  \emph{IFAC-PapersOnLine}, vol.~55, no.~24, pp. 266--272, 2022, 10th IFAC
  Symposium on Advances in Automotive Control AAC 2022.

\bibitem{weng2022leeged}
B.~Weng, G.~A. Castillo, W.~Zhang, and A.~Hereid, ``On safety testing,
  validation, and characterization with scenario-sampling: A case study of
  legged robots,'' in \emph{2022 IEEE/RSJ International Conference on
  Intelligent Robots and Systems (IROS)}.\hskip 1em plus 0.5em minus
  0.4em\relax IEEE, 2022, pp. 5179--5186.

\bibitem{corso2020survey}
A.~Corso, R.~Moss, M.~Koren, R.~Lee, and M.~Kochenderfer, ``A survey of
  algorithms for black-box safety validation of cyber-physical systems,''
  \emph{Journal of Artificial Intelligence Research}, vol.~72, pp. 377--428,
  2021.

\bibitem{gyllenhammar2020towards}
M.~Gyllenhammar, R.~Johansson, F.~Warg, D.~Chen, H.-M. Heyn, M.~Sanfridson,
  J.~S{\"o}derberg, A.~Thors{\'e}n, and S.~Ursing, ``Towards an operational
  design domain that supports the safety argumentation of an automated driving
  system,'' in \emph{10th European Congress on Embedded Real Time Systems (ERTS
  2020)}, 2020.

\bibitem{blanchini1999set}
F.~Blanchini, ``Set invariance in control,'' \emph{Automatica}, vol.~35,
  no.~11, pp. 1747--1767, 1999.

\bibitem{weng2021formal}
B.~Weng, L.~Capito, U.~Ozguner, and K.~Redmill, ``A formal characterization of
  black-box system safety performance with scenario sampling,'' \emph{IEEE
  Robotics and Automation Letters}, 2021.

\bibitem{sdq_tools}
B.~Weng, ``{SDQ} tools,'' \url{https://gitlab.com/Bobeye/sdq_tools}, 2022.

\bibitem{wang2018extracting}
W.~Wang and D.~Zhao, ``Extracting traffic primitives directly from
  naturalistically logged data for self-driving applications,'' \emph{IEEE
  Robotics and Automation Letters}, vol.~3, no.~2, pp. 1223--1229, 2018.

\bibitem{hauer2020clustering}
F.~Hauer, I.~Gerostathopoulos, T.~Schmidt, and A.~Pretschner, ``Clustering
  traffic scenarios using mental models as little as possible,'' in \emph{2020
  IEEE Intelligent Vehicles Symposium (IV)}.\hskip 1em plus 0.5em minus
  0.4em\relax IEEE, 2020, pp. 1007--1012.

\bibitem{scanlon2021waymo}
J.~M. Scanlon, K.~D. Kusano, T.~Daniel, C.~Alderson, A.~Ogle, and T.~Victor,
  ``Waymo simulated driving behavior in reconstructed fatal crashes within an
  autonomous vehicle operating domain,'' \emph{Accident Analysis \&
  Prevention}, vol. 163, p. 106454, 2021.

\bibitem{leung1989insights}
H.~K. Leung and L.~White, ``Insights into regression testing (software
  testing),'' in \emph{Proceedings. Conference on Software
  Maintenance-1989}.\hskip 1em plus 0.5em minus 0.4em\relax IEEE, 1989, pp.
  60--69.

\bibitem{yoo2012regression}
S.~Yoo and M.~Harman, ``Regression testing minimization, selection and
  prioritization: a survey,'' \emph{Software testing, verification and
  reliability}, vol.~22, no.~2, pp. 67--120, 2012.

\bibitem{onoma1998regression}
A.~K. Onoma, W.-T. Tsai, M.~Poonawala, and H.~Suganuma, ``Regression testing in
  an industrial environment,'' \emph{Communications of the ACM}, vol.~41,
  no.~5, pp. 81--86, 1998.

\bibitem{castillo2020hybrid}
G.~A. Castillo, B.~Weng, W.~Zhang, and A.~Hereid, ``Hybrid zero dynamics
  inspired feedback control policy design for 3d bipedal locomotion using
  reinforcement learning,'' in \emph{2020 IEEE International Conference on
  Robotics and Automation (ICRA)}.\hskip 1em plus 0.5em minus 0.4em\relax IEEE,
  2020, pp. 8746--8752.

\bibitem{Siekmann-RSS-20}
J.~Siekmann, S.~Valluri, J.~Dao, F.~Bermillo, H.~Duan, A.~Fern, and J.~Hurst,
  ``{Learning Memory-Based Control for Human-Scale Bipedal Locomotion},'' in
  \emph{Proceedings of Robotics: Science and Systems}, Corvalis, Oregon, USA,
  July 2020.

\bibitem{todorov2012mujoco}
E.~Todorov, T.~Erez, and Y.~Tassa, ``Mujoco: A physics engine for model-based
  control,'' in \emph{2012 IEEE/RSJ international conference on intelligent
  robots and systems}.\hskip 1em plus 0.5em minus 0.4em\relax IEEE, 2012, pp.
  5026--5033.

\bibitem{castillo2022reinforcement}
G.~A. Castillo, B.~Weng, W.~Zhang, and A.~Hereid, ``Reinforcement
  learning-based cascade motion policy design for robust 3d bipedal
  locomotion,'' \emph{IEEE Access}, vol.~10, pp. 20\,135--20\,148, 2022.

\bibitem{treiber2013traffic}
M.~Treiber and A.~Kesting, ``Traffic flow dynamics,'' \emph{Traffic Flow
  Dynamics: Data, Models and Simulation, Springer-Verlag Berlin Heidelberg},
  2013.

\bibitem{kesting2007general}
A.~Kesting, M.~Treiber, and D.~Helbing, ``General lane-changing model mobil for
  car-following models,'' \emph{Transportation Research Record}, vol. 1999,
  no.~1, pp. 86--94, 2007.

\bibitem{highway_env}
B.~Weng, ``{H}ighway{E}nv,'' \url{https://gitlab.com/Bobeye/highwayenv}, 2022.

\end{thebibliography}

\end{document}